\documentclass{article}
\usepackage{arxiv}
\usepackage [latin1]{inputenc}
\UseRawInputEncoding
\usepackage[utf8]{inputenx} 
\usepackage[T1]{fontenc}    
\usepackage{hyperref}       
\usepackage{url}            
\usepackage{booktabs}       
\usepackage{amsfonts}       
\usepackage{nicefrac}       
\usepackage{microtype}      
\usepackage{lipsum}
\usepackage{graphicx}
\usepackage{color}
\usepackage{tikz}
\usepackage{natbib}
\setcitestyle{authoryear,open={(},close={)}}
\usepackage{hyperref}
\usepackage{amsmath}
\usepackage{mathabx}
\usepackage{tikz}
\usepackage[export]{adjustbox}
\usepackage{multirow,array}
\usepackage{setspace}
\usepackage{graphics}
\usetikzlibrary{decorations.pathreplacing, positioning, backgrounds}
\title{Merging multiple input descriptors and supervisors in a deep neural network for tractogram filtering}
\date{}
\author{
  Daniel Jörgens \\
  Department of Biomedical Engineering and Health Systems\\
  KTH Royal Institute of Technology\\
  Stockholm, Sweden \\
  \texttt{danjorg@kth.se} \\
\And
 Pierre-Marc Jodoin\\
 Department of Computer Science\\
 Université de Sherbrooke\\
  Sherbrooke, Québec, Canada\\
  \texttt{pierre-marc.jodoin@usherbrooke.ca}
 \And Maxime Descoteaux\\
 Department of Computer Science\\
 Université de Sherbrooke\\
  Sherbrooke, Québec, Canada\\
   \texttt{M.Descoteaux@USherbrooke.ca}
 \And Rodrigo Moreno\\
  Department of Biomedical Engineering and Health Systems\\
  KTH Royal Institute of Technology\\
  Stockholm, Sweden \\
  \texttt{rodmore@kth.se}
  }
\begin{document}
\maketitle
\begin{abstract}
One of the main issues of the current tractography methods is their high false-positive rate. Tractogram filtering is an option to remove false-positive streamlines from tractography data in a post-processing step. In this paper, we train a deep neural network for filtering tractography data in which every streamline of a tractogram is classified as {\em plausible, implausible}, or {\em inconclusive}. For this, we use four different tractogram filtering strategies as supervisors: TractQuerier, RecobundlesX, TractSeg, and an anatomy-inspired filter. Their outputs are combined to obtain the classification labels for the streamlines. We assessed the importance of different types of information along the streamlines for performing this classification task, including the coordinates of the streamlines, diffusion data, landmarks, T1-weighted information, and a brain parcellation. We found that the streamline coordinates are the most relevant followed by the diffusion data in this particular classification task.
\end{abstract}
\keywords{Tractogram filtering \and Deep Learning \and Tractography \and Diffusion magnetic resonance imaging}
\section{Introduction}
Tractography is currently the method of choice to infer non-local structural connectivity information in the human brain \emph{in-vivo}. Following an estimate of local tissue orientation obtained from diffusion Magnetic Resonance Imaging (dMRI) data, tractography aims at producing streamlines that connect pairs of regions in the grey matter (GM). Every streamline is represented by a list of 3D coordinates. Ideally, a \textit{tractogram}, which is the set of streamlines computed through tractography, should reflect the underlying anatomy of white matter (WM) fibre bundles.
\par
Despite a long history of contributions to this field, current state-of-the-art tractography pipelines face a number of challenges \citep{Maier-Hein2017,Schilling2019a,Rheault2020CommonTractography}. These can result in a large number of anatomically implausible streamlines that do not reflect the underlying anatomy \citep{Maier-Hein2017} or are difficult to interpret in terms of known anatomy \citep{Petit2019}. Thus, careful consideration of which streamlines to select for subsequent analysis steps is crucial for drawing meaningful conclusions.
\par
A common strategy to address this problem is identifying and excluding those streamlines deemed implausible in a post-processing step after tractography. In the following, we will refer to this approach as \textit{tractogram filtering}. The resulting tractogram is expected to be more appropriate for its use in subsequent analyses after this step.
\par
Notice that defining the anatomical validity of streamlines is challenging since ground truth is unavailable. However, different methods have been proposed in the literature to obtain a subset of streamlines fulfilling specific conditions in a tractogram. These conditions can potentially model aspects of the anatomical plausibility of streamlines. 
\par
In our review \citep{Jorgens2021ChallengesFiltering}, we listed the most common approaches that have been used to define such rules. We identified five categories: methods using data fitting (e.g., \cite{Daducci2015,Smith2013,Pestilli2014,Hain2023}), methods using anatomical (e.g., \cite{Wasserthal2018,Wassermann2016a}) or geometrical priors (e.g., \cite{Petit2019,Astolfi2020TractogramLearning}), methods based on streamline bundle clustering (e.g., \cite{Garyfallidis2018,Zhang2018}) and methods that combine different approaches (e.g., \cite{Neher2018,Schiavi2020AInformation,Ocampo-Pineda2021HierarchicalTractography}). Machine learning has also been shown promising for tractogram filtering (e.g., \cite{Astolfi2020TractogramLearning,Legarreta2021FilteringFINTA,Hain2023}. It is important to remark that most of these methods were not originally proposed for tractogram filtering. However, they carry information on the plausibility of streamlines that can be used for that purpose.
\par
An important issue to consider is that the current methods are designed to  target either plausible or implausible streamlines, but not both sets. Thus, depending on the application, some filtering strategies are more suitable than others. To illustrate this dependency, we list some applications of tractography and their corresponding most appropriate filtering strategy: 
\begin{description}
    \item [Bundle-specific analysis:] in this case, the target anatomy is well-defined and known {\em a priori} (i.e., a particular fibre bundle). Thus, the focus of filtering must be to \textit{maximize the number of plausible streamlines} with respect to the definition of the targeted structure. Methods targeting plausible streamlines in specific bundles are the most appropriate for this application.
    \item [Whole-brain analysis:] whole-brain structural connectomics aims at finding global patterns of connections between different regions in the GM. In this case, the full set of targeted fibre bundles are challenging to define beforehand. Thus, the focus of tractogram filtering would be to \textit{reduce the number of implausible streamlines}. Methods targeting implausible streamlines are useful in this case.
    \item [Analysis in the presence of a pathology (e.g., a tumour):] tumours can change the anatomy of the WM. Filtering tractograms in this application is a delicate task since rules defining healthy conditions might no longer be applicable. For example, rules based on geometrical features of fibre bundles extracted from healthy subjects would not be adequate for bundles that are deformed due to an adjacent tumour. In this case, rules would optimally be independent of assumptions about healthy or pathological conditions. A combination of methods targeting {\em both plausible and implausible streamlines} is likely needed in this case.
    \item [Segmentation of new fibre bundles:] the current atlases of WM only contain well-defined fibre bundles. However, it could be of interest to investigate a bundle that is less well-defined, difficult to extract, or even undefined (e.g., the mentioned case of a tumour). A possible way to manually segment such bundles would be to remove streamlines that are related to a known set of bundles or implausible. As in the previous case, a combination of methods targeting {\em both plausible and implausible streamlines} is necessary for isolating  potential bundles of interest.
\end{description}
\par
As mentioned above, tractogram filtering can be an essential processing step for the following analyses. For example, \cite{Frigo2020DiffusionConnectomes}  reported that brain connectivity analyses could lead to different conclusions if the tractogram is not filtered. They also found out that the choice of filtering method has a direct influence on the results. We argue that combining different filtering methods could potentially increase the robustness and accuracy of connectivity analyses. 
\par
A standard approach in image analysis to combine the capabilities of different methods is to merge them in an ensemble fashion. As mentioned above, the current filtering methods target either plausible or implausible streamlines. Thus, basic ensemble strategies, such as majority voting, might not be optimal for combining approaches that target different types of streamlines. Moreover, this is also true for methods following the same strategy since they might be more or less specific in certain bundles. Notice that analysing specific descriptors of streamlines for which filtering methods disagree could potentially be used to improve filtering. Such an analysis would be possible by detecting streamlines for which the combination of individual methods is inconclusive.
\par
In this paper, we propose to group streamlines into three disjoint sets with the associated labels \emph{positive} (or plausible), \emph{negative} (or implausible) and \emph{inconclusive}. First, we define a mapping that combines the output of four different tractogram filtering methods to obtain the three labels. This combination is based on the analysis of the underlying rules in these methods. In a second step, we train a neural network that, using the four methods as supervisors, is able to perform the classification of streamlines using different input descriptors defined along with them. Finally, we assess the distribution of the three label classes, the performance of the network as well as the value of the different input descriptors in experiments on tractograms obtained from a subset of the Human Connectome Project (HCP) \citep{VanEssen2013TheOverview}.
\subsection*{Contributions}
In this paper, we have three main contributions. First, to the best of our knowledge, this is the first attempt to classify streamlines into three classes, including the new label \emph{inconclusive}. This label might be helpful in applications such as the exploration of weakly defined or undefined bundles of interest. Second, we propose a methodology for combining supervisors. This methodology is used to train a neural network that is able to reproduce both the individual supervisors as well as their combination. Finally, the structured assessment of the importance of different descriptors of the streamlines for performing the classification task gives valuable insights for future approaches to processing streamlines with neural networks.
\section{Materials and Methods}
\begin{figure}
    \centering
    \includegraphics[height=0.84\textheight]{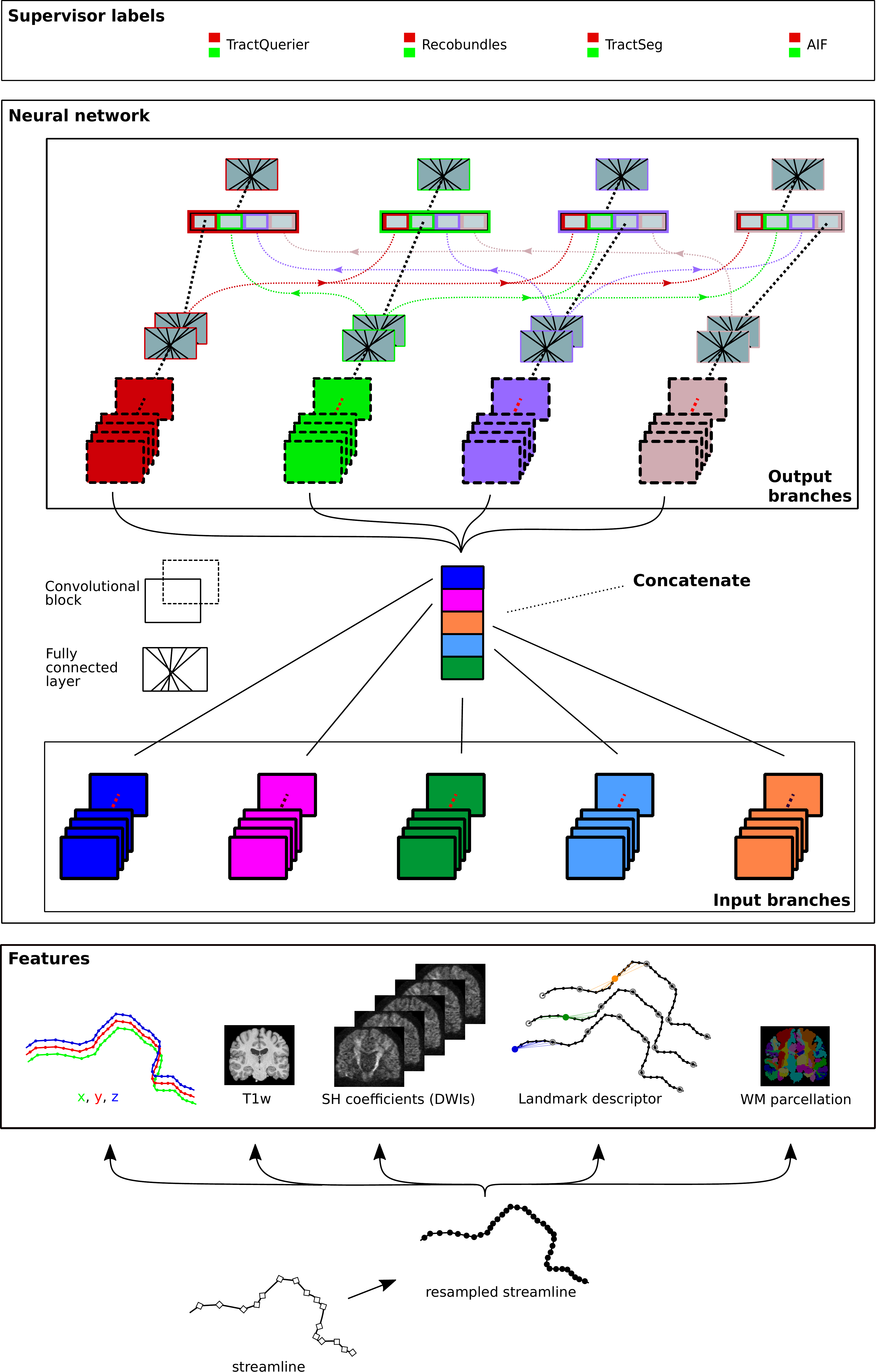}
    \caption{
        \footnotesize
        \setstretch{1.0}
        Overview of streamline classification approach. An individual streamline is resampled before five streamline- and image-based descriptors are extracted along its extent: the coordinates of the streamline (xyz), T1-weighted information, spherical harmonics coefficients of the diffusion data (SH), a landmark descriptor, and a brain parcellation. Each descriptor is processed in a separate input branch of the network. The labels of individual supervisor methods are predicted with separate output branches as well.
    }
    \label{fig:method_overview}
\end{figure}

In this section, we describe the input data and definition of the target labels, the chosen model architecture, the strategy for model training, as well as the performance metrics. We further outline the methodology based on which we aim to investigate the value of different input entities for the final prediction accuracy. Figure~\ref{fig:method_overview} gives an overview of the neural network approach for predicting the streamline labels. The goal of the neural network is to extract relevant features from the coordinates, T1w, dMRI, and brain parcellations along the streamline for predicting the labels given by four supervisors. The following subsections describe the different elements necessary to classify streamlines.
\subsection{Dataset}
\label{ssec:dataset}
Our dataset consists of whole-brain tractograms from 20 subjects in the HCP  \citep{VanEssen2013TheOverview}. Each tractogram has 10 million of streamlines obtained with Anatomically-Constrained Tractography (ACT) \citep{Smith2012} using the iFOD2 tracking algorithm \citep{TournierJ.-D.andCalamanteF.andConnelly2010} in MRtrix\footnote{\url{https://www.mrtrix.org}}. Note that these tractograms were also used in \cite{Wasserthal2018} as the basis for the dataset to train TractSeg. In addition to the tractograms, for each subject, we used the corresponding T1w image, dMRI data, and the FreeSurfer parcellation 'wmparc' precomputed in the HCP database. In this parcellation, WM locations within a distance of $5 \, mm$ to the GM are assigned with the label of the closest point in the GM \citep{Salat2009Age-associatedContrast}. The remaining WM voxels retain the same label distinguishing between the hemispheres.
\subsection{Label definition}
In this subsection, we describe our strategy for defining a plausibility label for each streamline. For this, we first employed four different strategies to obtain a binary label per streamline. In the following, we refer to these strategies as \emph{supervisors}. Subsequently, we shall specify how to combine the individual supervisor labels into the three classes: \emph{positive}, \emph{negative} and \emph{inconclusive}. Note that the used supervising strategies build upon approaches from literature, which have not been designed for binary classification of streamlines. Thus, we describe how we adapted them for such a purpose.
\subsubsection{Supervisor methods}
\label{ssec:supervisors}
We used the four supervisors described in the following paragraphs.
\paragraph{TractQuerier} The white matter query language (WMQL) and its python implementation \emph{TractQuerier} were proposed by \citet{Wassermann2016a} as a tool to structure the streamlines of tractograms according to rules based on brain anatomy. For example, it is possible with WMQL to define rules to select the streamlines that connect two GM regions and pass by a specific structure in the WM. Such regions are usually taken from a parcellation, e.g., \emph{wmparc} from FreeSurfer. Streamlines fulfilling all  defined constraints are labelled as potential members of the corresponding fibre bundle of interest.
\par
In this paper, we use the set of rules proposed in \cite{Wasserthal2018}\footnote{\url{https://github.com/MIC-DKFZ/TractSeg/blob/0a1947c5cab84afbbd8fa8f76d4e82d126ba0640/resources/WMQL_Queries.qry}} for defining 72 different WM bundles. After applying TractQuerier, we assign a positive label to all streamlines fulfilling the constraints for at least one of these 72 bundles defined in WMQL. Notice that streamlines not matching any of the bundle definitions could either belong to a bundle that is not part of the definitions or reflect an anatomically valid structure. Thus, the rules of TractQuerier are necessary conditions for a streamline of a particular bundle to be plausible but are not sufficient for acceptance with high certainty. For example, it is possible to find streamlines that comply with the specific criteria checked by TractQuerier, but that are locally deviating from an anatomically interpretable shape (e.g., streamlines with loops are unlikely to represent the underlying anatomy of WM).
\paragraph{RecobundlesX } The aim of \emph{Recobundles} \citep{Garyfallidis2018} is to extract streamlines from a whole-brain tractogram that comply with the definitions of a particular set of WM bundles. In contrast to TractQuerier, these definitions are not explicitly stated as rules but implicitly encoded in an atlas of streamlines. Through multiple steps of streamline clustering and registration, those streamlines that are likely to match a specific bundle definition in the atlas are extracted. The main assumption of Recobundles is that the shape of streamlines in the same bundle is similar in the atlas and the tractogram of a particular subject.
\par
In this paper, we assign a positive label to streamlines recognised as part of a bundle in the used atlas. Similar to TractQuerier, streamlines not extracted with this approach can be either implausible or be part of a bundle that is not part of the atlas. The employed atlas consists of two subjects with 29 defined bundles each\footnote{\url{https://zenodo.org/record/4104300\#.YP_cIFNKhQM}}. For bundle segmentation, we use a multi-atlas strategy termed \emph{RecobundlesX} \citep{Rheault2020AtlasRecobundlesX} \footnote{\url{https://github.com/scilus/scilpy/blob/master/scripts/scil_recognize_multi_bundles.py}}.
\paragraph{TractSeg} \emph{TractSeg} \citep{Wasserthal2018} is a neural network-based approach that uses the fibre orientation distribution (FOD) peaks extracted from dMRI data to directly compute the segmentation masks for 72 different WM bundles. Moreover, it is possible to compute a pair of endpoint regions for each bundle with this method. In the same way as for TractQuerier and RecobundlesX, we can employ the bundle definitions in TractSeg to define a binary streamline classification.
\par
In order to obtain the binary TractSeg labels, we use the segmentation masks of the 72 bundles and their corresponding endpoint regions to define constraints for streamlines belonging to a particular one of them. More specifically, let $B$ be a bundle of streamlines, $s \equiv (s_0, ..., s_N)$ a streamline represented by $N$ sampling points, and $M_B, E_B^0, E_B^1$ the segmentation mask of the bundle and the endpoint masks of $B$, respectively. Then, streamlines in $B$ must comply with the following rule:
\begin{equation}
    s \in B \Rightarrow s_i \in M_B \; \forall_{s_i \in s} \land ((s_0 \in E_B^0 \land s_N \in E_B^1) \lor (s_N \in E_B^0 \land s_0 \in E_B^1)).
    \label{eq:ts_condition}
\end{equation}
As before, we assume streamlines complying with this rule are plausible and are assigned a positive label. Likewise, a streamline that is not selected may either belong to a bundle that TractSeg does not target or be in general implausible. Additionally, similarly to TractQuerier, Eq.~(\ref{eq:ts_condition}) is a necessary condition for a plausible streamline in one of the defined bundles.
\paragraph{Anatomy-inspired filtering (AIF)} Employing simple rules inspired by anatomical constraints has been investigated as an approach to obtain cleaner tractograms \citep{CoteMarc-AlexandreandGaryfallidisEleftheriosandLarochelleHugoandDescoteaux2015CleaningClustering,Smith2012,Girard2014TowardsBiases,Zhang2018MappingConnectomes,Petit2019}. Based on this observation, we use a fourth supervising strategy, which implements a set of these ideas. With this strategy, we aim to remove
\begin{itemize}
    \item streamlines with a loop of more than $360^\circ$ \citep{CoteMarc-AlexandreandGaryfallidisEleftheriosandLarochelleHugoandDescoteaux2015CleaningClustering};
    \item streamlines ending within the deep white matter or near the ventricles \citep{Petit2019}.
\end{itemize}
Any streamline identified through these rules is assigned a negative label. Notice that, unlike the other supervisors, this approach is tailored to remove implausible streamlines. Therefore the positive label can be seen as a necessary but not a sufficient condition for plausible streamlines. This means that a streamline complying with the defined rules can still be wrong. In the following, we will refer to this supervisor as anatomy-inspired filtering (AIF).
%
%
Table~\ref{tab:supervisors} summarizes the four used supervisors and details the interpretation of their binary labels. 
\begin{table}[ht]
    \footnotesize
    \centering
    \caption{Interpretation of the classification output of each supervisor.}
    \begin{tabular}{@{}l@{\hskip 2mm}p{5.3cm}@{\hskip 2mm}p{5.3cm}@{}}
        \toprule
        Supervisor & Positive label & Negative label \\
        \midrule
        TractQuerier & Plausible: candidate for a defined bundle & Inconclusive: implausible or member of an undefined bundle \\
        RecobundlesX & Plausible: member of a defined bundle & Inconclusive: implausible or member of an undefined bundle \\
        TractSeg & Plausible: candidate for a defined bundle & Inconclusive: implausible or member of an undefined bundle \\
        AIF & Inconclusive: more information is needed for assessing plausibility & Implausible \\
        \bottomrule
    \end{tabular}
    \label{tab:supervisors}
\end{table}
\subsubsection{Classification of streamlines}
\label{sssec:three_class_problem}
Based on the four supervisor labels defined in the preceding subsection, our goal is to divide the streamlines in a tractogram into three disjoint classes: positive (POS), negative (NEG), unknown/inconclusive (U). As mentioned above, it is not obvious how to combine the labels from these four methods. In this section, we propose a methodology consisting of two steps in order to achieve this.
\par
First, we consider all combinations of the four binary labels, resulting in $2^4=16$ classes.  Table \ref{tab:composition_classes} provides an interpretation of the samples falling in each respective class. While we expect many of these classes to be found in the data, others are expected to be uncommon (borderline classes in Table \ref{tab:composition_classes}). The occurrence of the latter is expected to be due to inaccuracies in processing operations. 
\begin{table}
    \footnotesize
    \def\widthlastcolumn{2.4cm}
    \centering
    \caption{Interpretation of combined supervisor labels. The sixteen possible combinations of the individual supervisor labels are grouped with respect to their proposed interpretation. \textbf{TQ}: TractQuerier, \textbf{RBX}: RecobundlesX, \textbf{TS}: TractSeg, \textbf{AIF}: Anatomy-inspired filtering, \textbf{p}: positive, \textbf{n}: negative.}
    \label{tab:composition_classes}
    \begin{tabular}{@{}c@{\hskip 2mm}c@{\hskip 2mm}c@{\hskip 2mm}c@{\hskip 2mm}p{7.4cm}@{\hskip 3mm}p{\widthlastcolumn}@{}}
        \toprule
        TQ & RBX & TS & AIF & Description & Interpretation \\
        \midrule
        p & p & p & p & Plausible streamline & Likely plausible \\
        p & n & p & p & Streamline might not be part of the RBX atlas & Likely plausible \\
        p & p & n & p & TS mask might be too strict and locally violated  & Likely plausible \\
        n & p & n & p & TQ and TS masks might be locally violated  (e.g., at endpoints) & Likely plausible \\
        n & p & p & p & TQ can disagree with TS due to small misalignments & Likely plausible \\
        \midrule
        n & n & n & n & Implausible streamline &Likely implausible \\
        p & n & n & n & The shape of the streamline might be wrong & Likely implausible \\
        p & n & p & n & Streamline might have loops within the bundle mask of TS & Likely implausible \\
        \midrule
        n & n & n & p & Streamline is not captured by neither the RBX atlas nor TQ and TS masks & Inconclusive \\
        p & n & n & p & The implemented rules of TQ are less specific than RBX and TS, more information is required to assess plausibility & Inconclusive \\
        \midrule
        n & n & p & p & TQ can disagree with TS due to small misalignments. The streamline might not be in the RBX atlas & \parbox[t]{\widthlastcolumn}{Likely plausible \\ (borderline)} \\
        p & p & n & n & TS and AIF masks might be locally violated or the shape can be wrong &\parbox[t]{\widthlastcolumn}{Likely implausible\\ (borderline)} \\
        p & p & p & n & AIF masks might be locally violated or the shape might be wrong & \parbox[t]{\widthlastcolumn}{Likely implausible\\ (borderline)} \\
        n & p & n & n & RBX tolerates violation of TQ, TS, and AIF masks (e.g., at endpoints) or the shape is wrong & \parbox[t]{\widthlastcolumn}{Likely implausible\\ (borderline)} \\
        n & n & p & n & TQ can disagree with TS due to small misalignments, but the shape might be wrong & \parbox[t]{\widthlastcolumn}{Likely implausible \\ (borderline)} \\
        n & p & p & n & TQ can disagree with TS due to small misalignments, the streamline is in the RBX atlas, but the shape might be wrong & \parbox[t]{\widthlastcolumn}{Likely implausible \\ (borderline)} \\
        \bottomrule
    \end{tabular}
\end{table}
\par
In a second step, based on the analysis of Table \ref{tab:composition_classes}, these sixteen classes are each assigned to one of the three target labels POS, NEG, or U. These classes are defined as follows:
\paragraph{POS - positive (plausible)} Both RecobundlesX and TractSeg have a high specificity in their positive label. Therefore, we regard combinations in which AIF gives a positive label and RecobundlesX, or TractSeg do as well as a positive (POS) class.
\paragraph{NEG - negative (implausible)} Since the negative label of AIF has a high specificity, it overrules all other labels in the ensembling step. Therefore, whenever AIF provides a negative label, the combination of binary labels is mapped to the class NEG.
\paragraph{U - unknown/inconclusive} This group contains all cases where AIF gives a positive label, but both RecobundlesX and TractSeg give negative ones. The TractQuerier label is not enough to distinguish between positive and negative cases due to the lower specificity of its positive label.
\begin{table}
    \footnotesize
    \centering
    \caption{Classification of streamlines into POS, NEG, and U. Each item listed in the second column shows the combination of the four supervisor labels that are mapped to the particular three-class label and corresponds to a row in Table~\ref{tab:composition_classes}. The order of the supervisor labels is TQ, RB, TS, AIF. \textbf{TQ}: TractQuerier, \textbf{RBX}: RecobundlesX, \textbf{TS}: TractSeg, \textbf{AIF}: Anatomy-inspired filtering, \textbf{p}: positive, \textbf{n}: negative.}
    \label{tab:three_class_problem}
    \begin{tabular}{@{}l@{\hskip 3mm}p{6cm}@{\hskip 3mm}p{6cm}@{}}
        \toprule
        Label & Composition (TQ, RBX, TS, AIF) & Criteria \\
        \midrule
        POS & pppp, pnpp, ppnp, npnp, nnpp, nppp  & AIF label is positive and RB and/or TS are positive. \\
        NEG & nnnn, pnnn, pnpn, ppnn, pppn, npnn, nnpn, nppn & AIF label is negative. \\
        U & nnnp, pnnp  & AIF label is positive, and RB and TS are negative. \\
        \bottomrule
    \end{tabular}
\end{table}
Table \ref{tab:three_class_problem} summarises the mapping between the individual labels to the aforementioned three labels. Although borderline cases are expected to cover only a small amount of streamlines, these are either likely plausible or implausible depending on the individual labels and thus can be added to the POS or NEG label.
\subsection{Neural network architecture}
The classifier is a star-shaped neural network as depicted in Fig. \ref{fig:method_overview}.  As shown, it is composed of five input branches whose features are concatenated into one block before passing them on to four output branches. Each input branch is associated with one of the input streamline descriptors, and each output branch with one of the supervisor methods. The input descriptors are explained in detail in  Subsect.~\ref{ssec:input_features}.
\par
At the end of the input branches, their features are concatenated along the channel dimension before forwarding them to the output branches. In this way, early features in the input branches might be shared for predicting different supervisor labels, which is a standard procedure in multi-task problems like the one at hand. In the same way, also the features of the second fully connected (FC) layer of each output branch are shared among all branches and concatenated before passing them on to the last two FC layers. Note that these connections are one-way, i.e. the features of a particular branch are forwarded to all other branches, but during backpropagation, only the gradients computed for the same output branch are considered. Through this architecture, the output branches remain independent from each other in the optimisation phase, but at the same time, their late features can facilitate the predictions of other supervisor labels as well.
\par
All branches consist of series of convolutional blocks. Such a block comprises a one-dimensional convolution, followed by batch normalization, a ReLU activation function and optional pooling. In our experiments, each input branch was composed of 12 and each output branch of 4 convolutional blocks. The features of the last convolutional block in each output branch are flattened and forwarded to a series of four FC layers with ReLU activation. After the concatenation of the features from all output branches described in the previous paragraph, another FC layer is employed. The final prediction is obtained from 2-neuron softmax outputs. A detailed summary of the architecture is provided in Appendix \ref{app:network}. The implementation of the neural network was done using the \textit{PyTorch} framework  \citep{Paszke2019PyTorch:Library}. The star-like architecture is useful for investigating both the value of different input sources on the final streamline classification and the capability of the network to predict the different supervisor labels.
\subsection{Input descriptors}
\label{ssec:input_features}
Each streamline depicts an individual sample to be classified. We process the streamlines as follows. First, each streamline is resampled to a constant step size of $1 \, mm$. After that, the number of sampling points is restricted to the first $N=100$. In case the total number of steps of the streamline is less than $N$, the remaining positions are filled with zeros. After this, we extract the five input descriptors detailed in the next paragraphs.
\paragraph{Streamline coordinates} We use the three coordinates of every sample point of the resampled streamline as data channels. The coordinates are normalized to the range [-1,1]. This results in a $3 \times N$-shaped data array to input to the network.
\paragraph{Landmark-based shape descriptor} 
Using the coordinates of the streamline alone makes the classifier susceptible to translations and rotations in the data. Thus, it is interesting to capture the streamline shape in a way that is invariant to rotations and translations. Inspired by the work of \cite{NgattaiLam2018}, we add a landmark-based shape descriptor from each streamline that we used as an input to the network. This shape descriptor, $\mathrm{LM}[s]$, is computed as follows. First, we set a fixed number of landmarks $K$ that are equidistantly distributed along each streamline. Then, $\mathrm{LM}[s]$ is computed as the distance between the landmarks to all points on the streamline. More formally, the descriptor for the $k$-th landmark is given by
\begin{equation}
    \mathrm{LM}_k(s) := \left\{||l_k-s_i||  \forall s_i \in s \right\},
\end{equation}
where $l_k$ and $s_i$ are the coordinates of the $k$-th landmark and the $i$-th sample point, respectively. Thus, $\mathrm{LM}[s]$ is a $K \times N$-shaped data array. In the experiments of Section~\ref{sec:results}, we used $K=20$. 
\par
Note that the landmarks are defined locally, i.e., within each streamline. As opposed to the proposal in \cite{NgattaiLam2018}, this design makes the descriptor invariant to translations and rotations that might affect the coordinates of the streamlines. On the other hand, these invariance properties come at the cost of losing information about the streamline captured in this feature compared to the full set of streamline coordinates.
\paragraph{Spherical harmonics transform of the diffusion signal} Since streamlines should optimally comply with the acquired data, we include a feature derived from the diffusion signal in the network input. For this, we fit a spherical harmonics (SH) expansion of order 4 to the diffusion signal for each $b$-shell ($b \in \{1000,3000\}$) individually, obtaining 28 SH coefficients per voxel and shell. After concatenating the two images of SH coefficients for both shells along the dimension of the SH coefficients, we linearly interpolate the volume at the sampling points of each streamline, which results in a $56 \times N$-shaped data array. This array is used as the input of the network.
\paragraph{T1-weighted data} Local information of the type of tissue (GM, WM, or CSF) might be useful for disregarding erroneous streamlines. Thus, we add a feature derived from the T1-weighted (T1w) image values along the streamlines as another classifier input. After normalisation with respect to the minimum and maximum gray-scale value, the T1w image of each subject is linearly interpolated at the sampling points of a streamline. This results in a $1 \times N$-shaped data array.
\paragraph{White matter parcellation} Streamlines belonging to the same bundle are expected to follow a similar path throughout the brain. Therefore, the information of which WM regions are traversed along the course of a streamline might provide useful, subject-specific information to classify streamlines. 
\par
In order to make use of the categorical information of the parcellation as an input to our network, we utilise a `one-hot' encoding. With the purpose of modeling the inclusion of different sample points of the streamline in different WM regions, we generate a matrix $W$ per streamline such that
\begin{equation}
    W_{ij} = \left\{
    \begin{array}{ll}
        1 & s_j \in R_i \\
        0 & otherwise
    \end{array}
    \right.,  
\end{equation}
where $R_i$ is the $i$-th region in the parcellation and $s_j$ is a sample point. This matrix has a size of $L\times N$, with $L$ being the number of regions in the parcellation. As mentioned above, we used the `wmparc' parcellation generated by FreeSurfer.
\subsection{Model training}
\label{ssec:model_training}
For training the neural network, we use \textit{Adam} optimiser \citep{Kingma2015Adam:Optimization} with a learning rate $3\cdot\mathrm{10}^{-5}$ for minimising the sum of cross-entropy losses over all four model output branches. We restrict the number of epochs to a maximum of 250.
\par
The available data of 20 subjects is divided at the subject level into a training set (for computing the gradients in the loss optimisation), a validation set (for estimating the ability of the model to generalise to independent data during training), and a test set (to assess the performance of the final model configuration on unseen/independent data after completion of the training phase). For each trained model, we randomly divide all 20 subjects into these sets with a ratio of 60\%  for training, 25\% for validation, and  15\%  for testing, i.e., twelve, five, and three subjects, respectively.
\par
Since our dataset has a large number of samples (12 subjects with 10 million samples each, resulting in 120 million samples for training), we select a suitable subset of samples for optimisation within each epoch. In our experiments, we picked a random set of 10,000 samples from the training set to optimise the model and 4,000 samples from the validation set to estimate the quality of the current set of optimised model parameters. In order to balance both the occurrence of the different subject and labels  considering the  sixteen different classes, we apply the following hierarchical sampling strategy:
\begin{itemize}
    \item First, a subject id ($T_i$) is picked from the pool of all training/validation subjects weighted by the number of streamlines available for each subject. This ensures that the distribution of subject ids is the same as if streamlines were picked from the pool of all available streamlines.
    \item Second, a label id ($L_j$) is picked from the pool of all sixteen composition labels with equal probabilities for all labels. This ensures that the distribution of labels for each of the supervisor methods is uniform as well.
    \item Finally, a streamline id is picked from the pool of all streamlines of subject $T_i$ with label $L_j$.
\end{itemize}
While this sampling strategy is meaningful in our case, it also has potential shortcomings. On the one hand, in case the subjects have very unbalanced label classes and the number of drawn samples in each of the subjects is much larger than the number of samples for the smallest label class, this sampling strategy might result into many repeated samples in the small label class.
Secondly, duplicates within one epoch can generally not be avoided with this strategy due to the random sampling approach. 
We assume that these potential issues can have a minor effect on training due to the large number of samples per subject and labels in each subject. 
\subsection{Measurements for evaluation}
For the assessment of performance at training and evaluation time, we use several metrics. In particular, the individual binary cross-entropy, accuracy, precision and recall of each output branch is used to assess the progress in the optimisation of the individual supervisors, as well as their mean. All test metrics are estimated through averages of the individual experiment outcomes for five random realisations of the mentioned data splitting into train, validation and test sets.
\subsection{Ablation study}
\label{ssec:value_of_inputs}
We build our model on the data of five different streamline inputs. Thus, we systematically assessed the value of each of them for the final prediction performance of the model. For this purpose, we retrain our model with the same set of hyperparameters while replacing the information of one or more input modalities (i.e., all input data to one or several of the input branches) with standard normal distributed random values. By this approach, we effectively remove any information given to the particular input branches while maintaining the number of trainable parameters in the model. Thus, the capacity of the model remains unchanged.
\section{Results} \label{sec:results}
\subsection{Label creation}
\begin{figure}
    \centering
    \includegraphics[width=\textwidth]{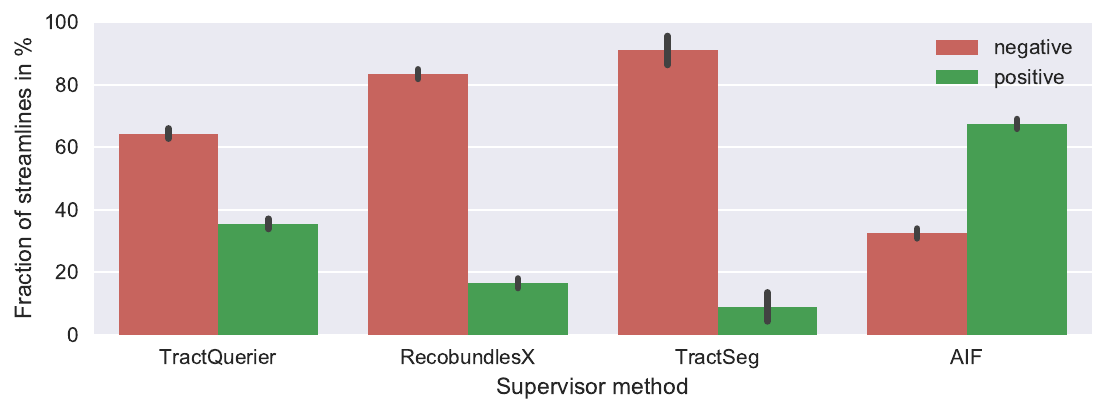}
    \caption{Average number of streamlines per label and supervisor method. The number of streamlines for the positive and negative labels obtained from each of the supervisor methods is averaged over all 20 subjects. The error bars indicate the standard deviation for the particular groups.}
    \label{fig:supervisor_stats_4}
\end{figure}
\begin{figure}
    \centering
    \includegraphics[width=\textwidth]{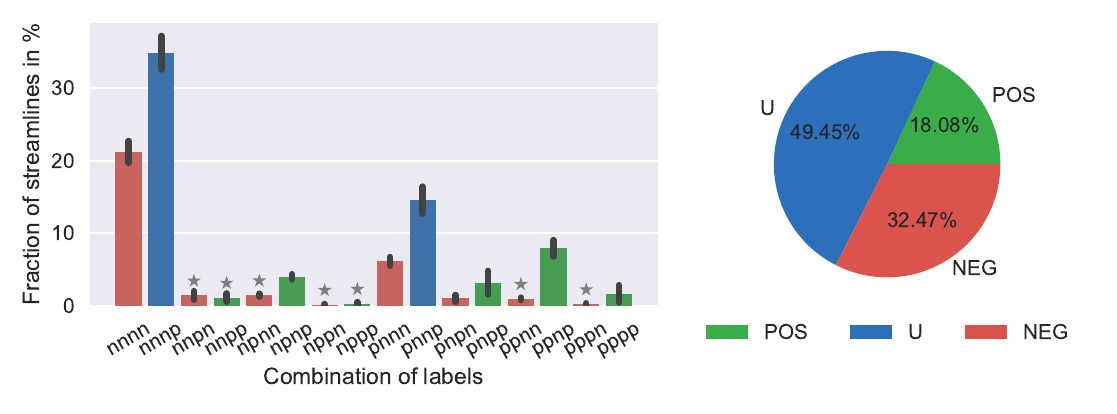}
    \caption{Average number of streamlines per combination of supervisor labels and in the derived three-class problem (POS, NEG, U). \textbf{Left:}  percentage of streamlines for all sixteen combination of individual supervisor labels (TQ, RBX, TS, AIF)  averaged over all 20 subjects. Error bars indicate the standard deviation for every particular class. Classes listed as borderline in Table~\ref{tab:composition_classes} are annotated with a star marker. \textbf{Right:} Further grouping of the labels according to Table~\ref{tab:three_class_problem} results in the distribution of the three classes \textit{POS} (\raisebox{0.6pt}{\protect\tikz{\protect\draw[green!70!black,scale=0.3,fill] (0,0) rectangle (0.5,0.5);}}), \textit{NEG} (\raisebox{0.6pt}{\protect\tikz{\protect\draw[red!70!white,scale=0.3,fill] (0,0) rectangle (0.5,0.5);}}) and \textit{U} (\raisebox{0.6pt}{\protect\tikz{\protect\draw[blue!70!white,scale=0.3,fill] (0,0) rectangle (0.5,0.5);}}) depicted in the pie chart. Note that the coloring of the combinations of labels in the bar chart on the left corresponds to the grouping of the three-class problem shown on the right.}
    \label{fig:label_balance_4}
\end{figure}
\begin{figure}
    \footnotesize
    \centering
    \def\ydist{0cm}
    \begin{tikzpicture}[background rectangle/.style={fill=black}, show background rectangle, node distance=0mm and 0mm]
        %
        \node (nnnn) {\includegraphics[trim=4cm 8cm 4cm 7.5cm,clip,width=3.cm]{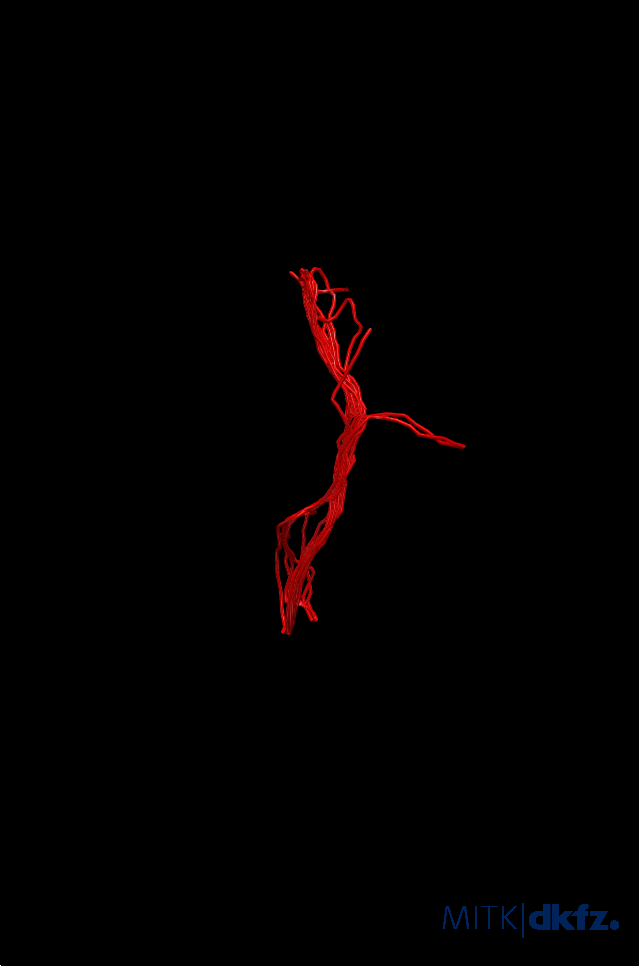}};
        \node[right=of nnnn] (nnnp) {\includegraphics[trim=4cm 8cm 4cm 7.5cm,clip,width=3.cm]{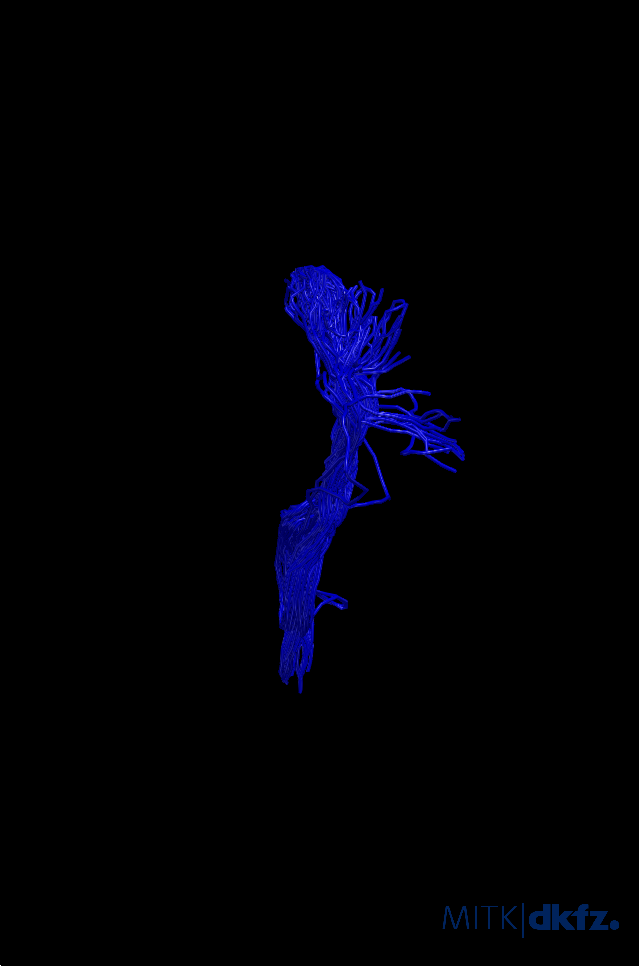}};
        \node[right=of nnnp] (nnpn) {\includegraphics[trim=1cm 5cm 1cm 4.5cm,clip,width=3.cm]{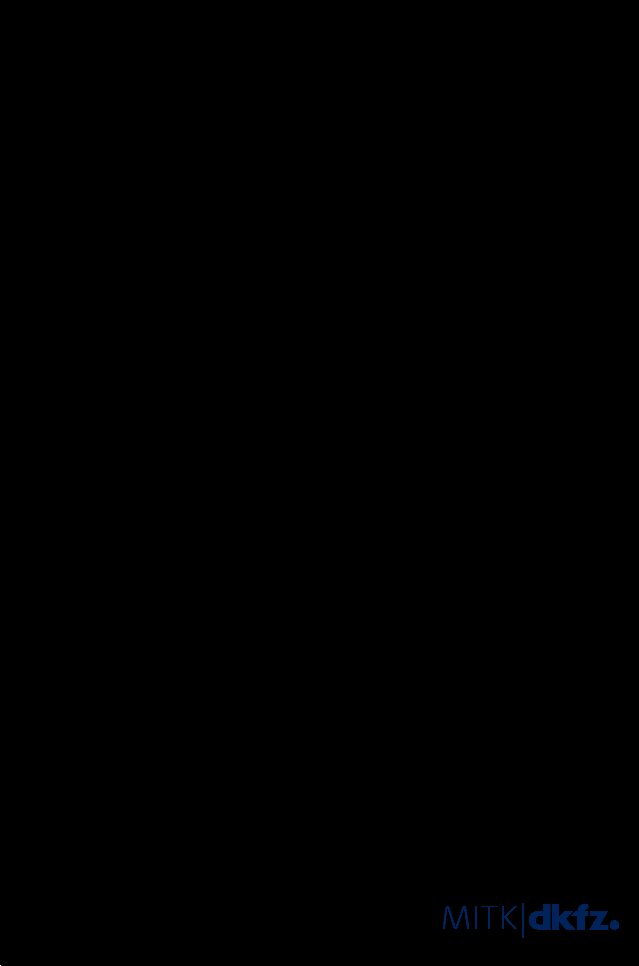}};
        \node[right=of nnpn] (nnpp) {\includegraphics[trim=1cm 5cm 1cm 4.5cm,clip,width=3.cm]{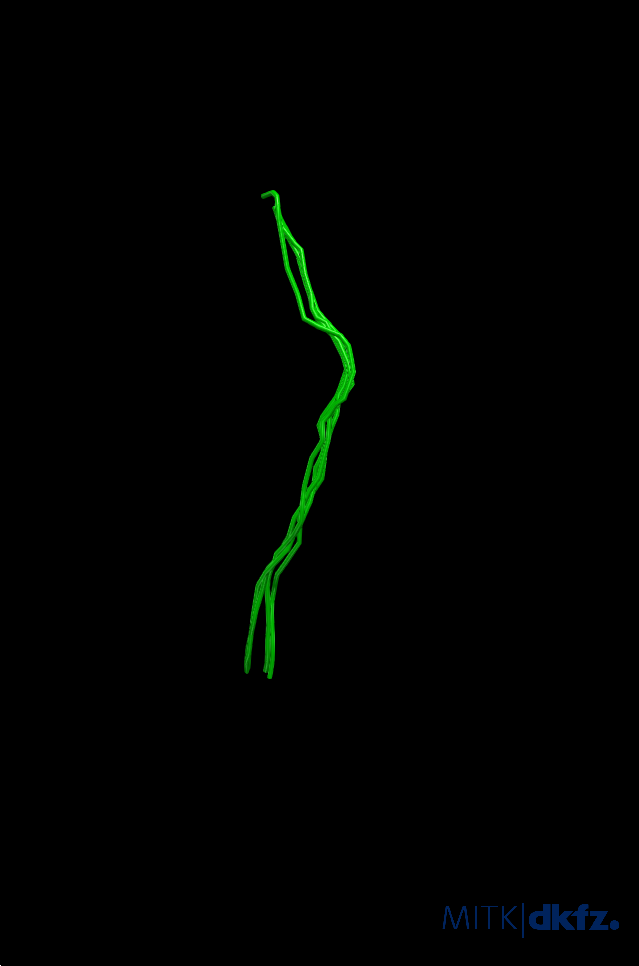}};
        %
        \node[below=of nnnn] (npnn) {\includegraphics[trim=1cm 5cm 1cm 4.5cm,clip,width=3.cm]{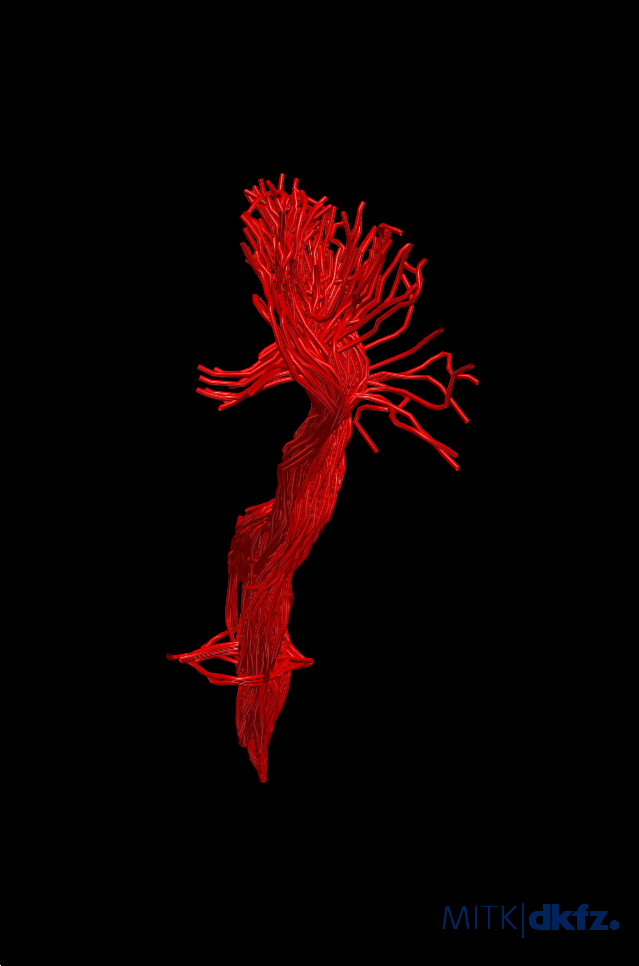}};
        \node[right=of npnn] (npnp) {\includegraphics[trim=1cm 5cm 1cm 4.5cm,clip,width=3.cm]{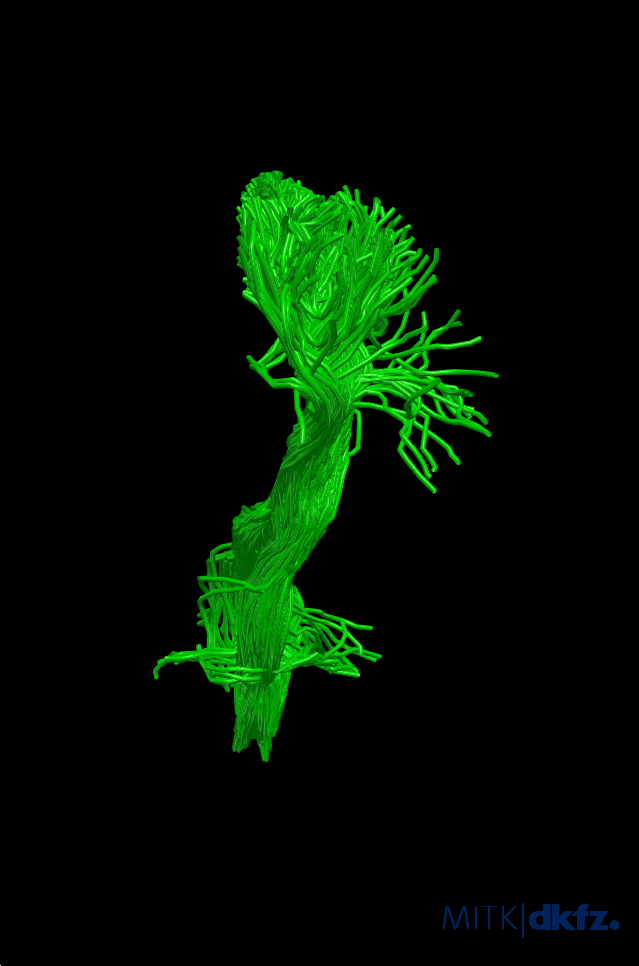}};
        \node[right=of npnp] (nppn) {\includegraphics[trim=1cm 5cm 1cm 4.5cm,clip,width=3.cm]{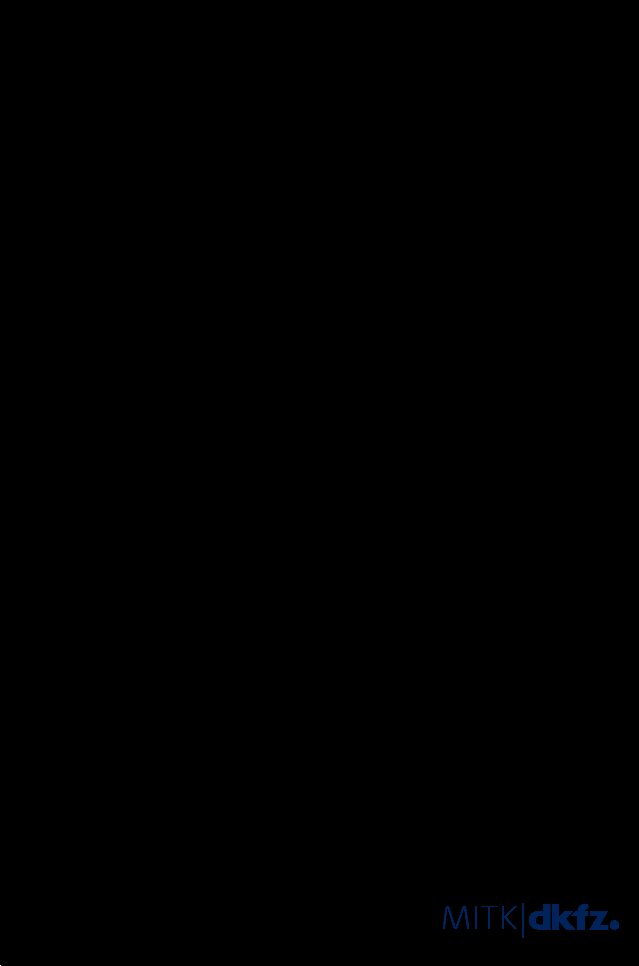}};
        \node[right=of nppn] (nppp) {\includegraphics[trim=1cm 5cm 1cm 4.5cm,clip,width=3.cm]{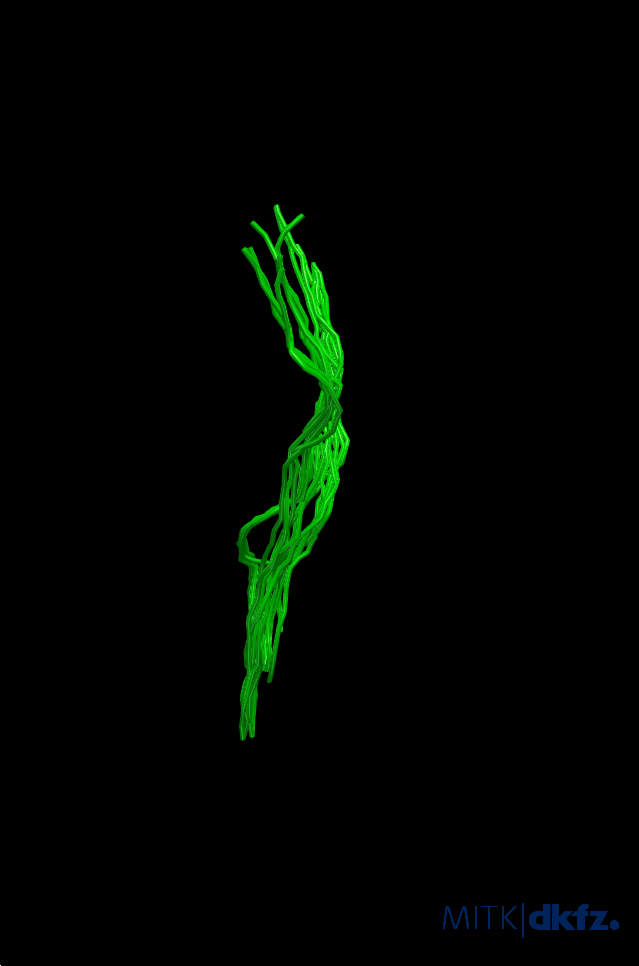}};
        %
        \node[below=of npnn] (pnnn) {\includegraphics[trim=1cm 5cm 1cm 4.5cm,clip,width=3.cm]{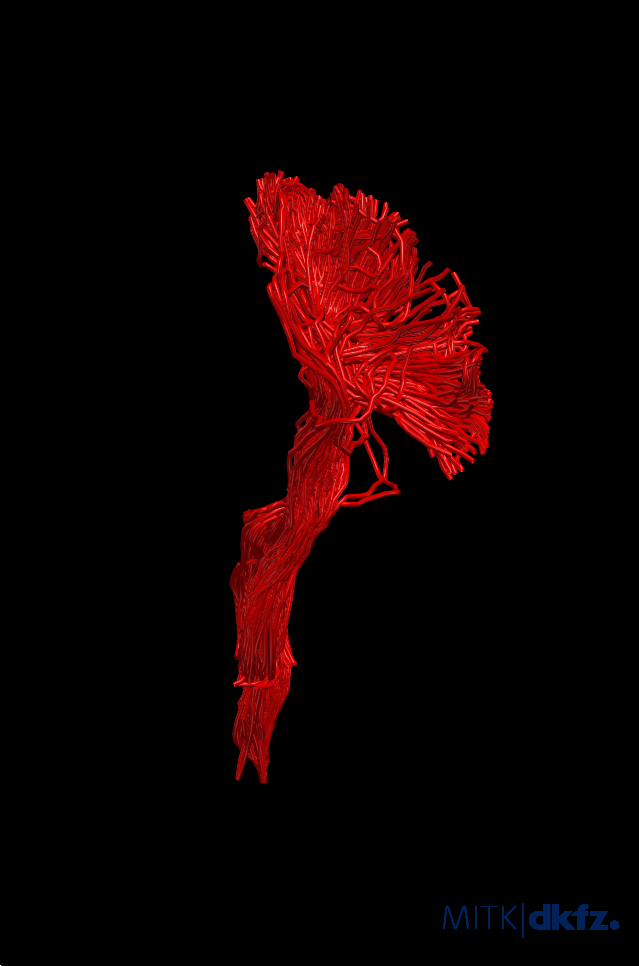}};
        \node[right=of pnnn] (pnnp) {\includegraphics[trim=1cm 5cm 1cm 4.5cm,clip,width=3.cm]{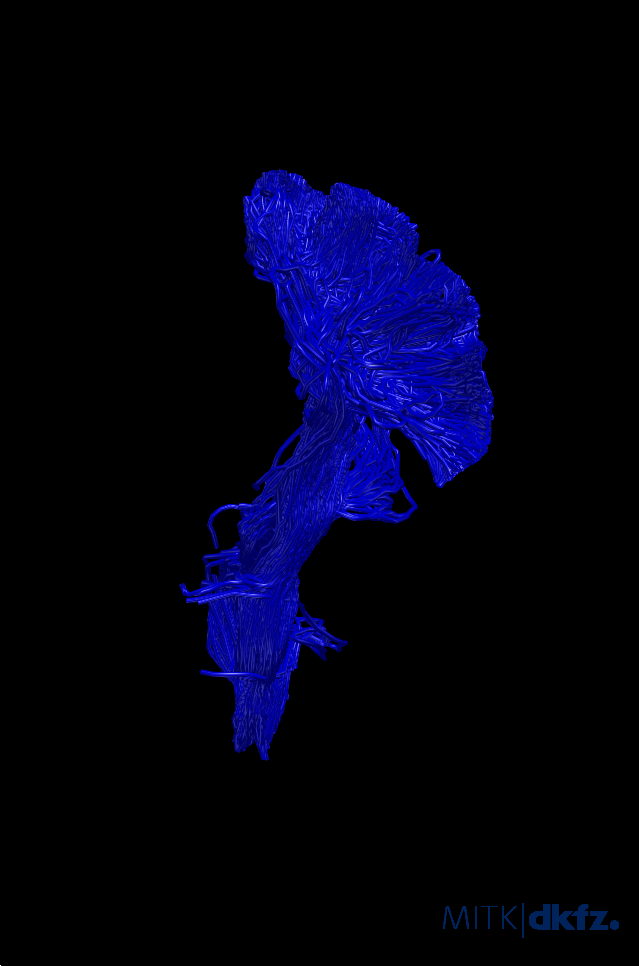}};
        \node[right=of pnnp] (pnpn) {\includegraphics[trim=1cm 5cm 1cm 4.5cm,clip,width=3.cm]{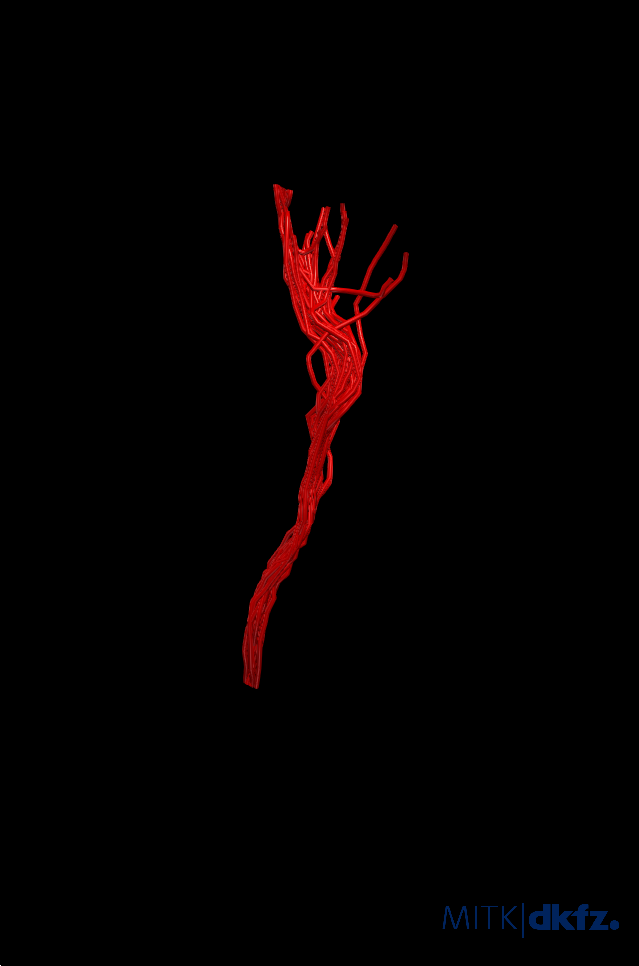}};
        \node[right=of pnpn] (pnpp) {\includegraphics[trim=1cm 5cm 1cm 4.5cm,clip,width=3.cm]{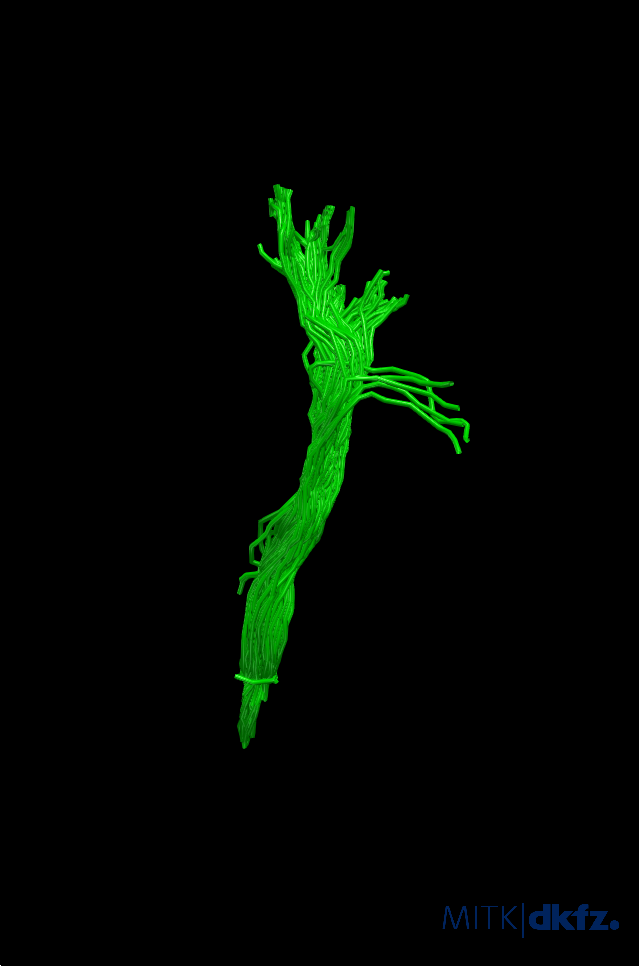}};
        %
        \node[below=of pnnn] (ppnn) {\includegraphics[trim=1cm 5cm 1cm 4.5cm,clip,width=3.cm]{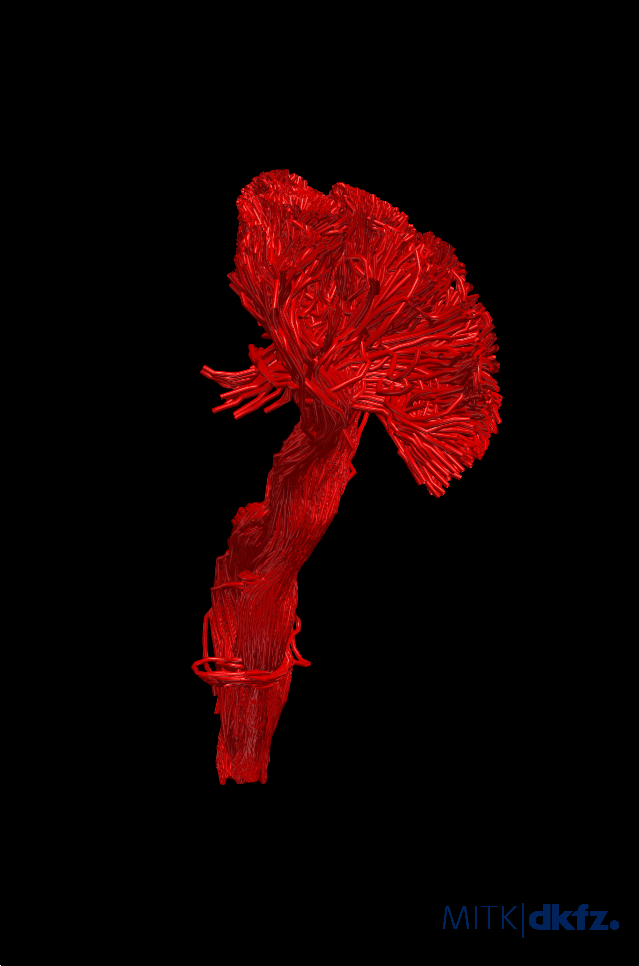}};
        \node[right=of ppnn] (ppnp) {\includegraphics[trim=1cm 5cm 1cm 4.5cm,clip,width=3.cm]{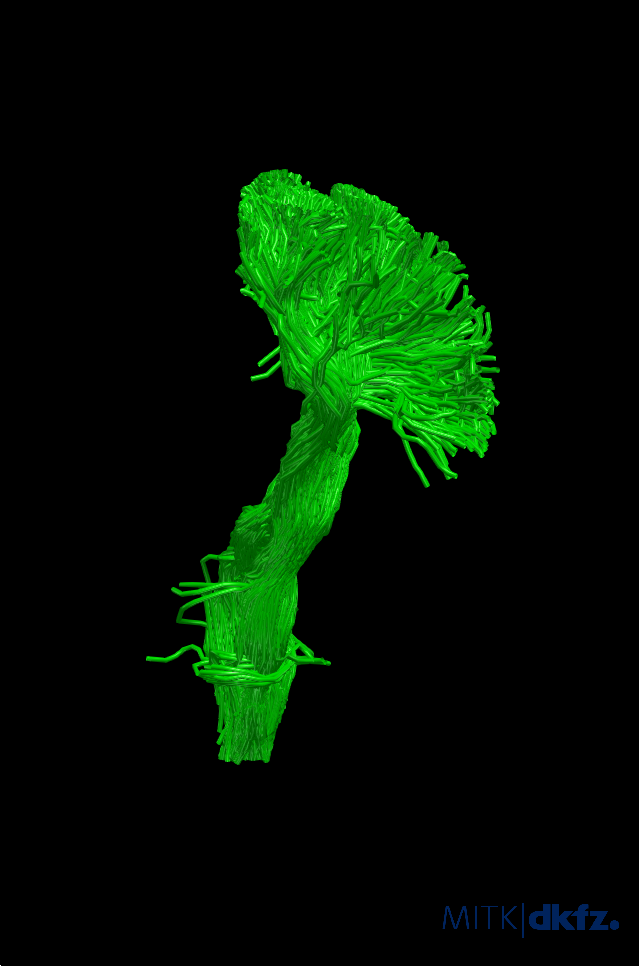}};
        \node[right=of ppnp] (pppn) {\includegraphics[trim=1cm 5cm 1cm 4.5cm,clip,width=3.cm]{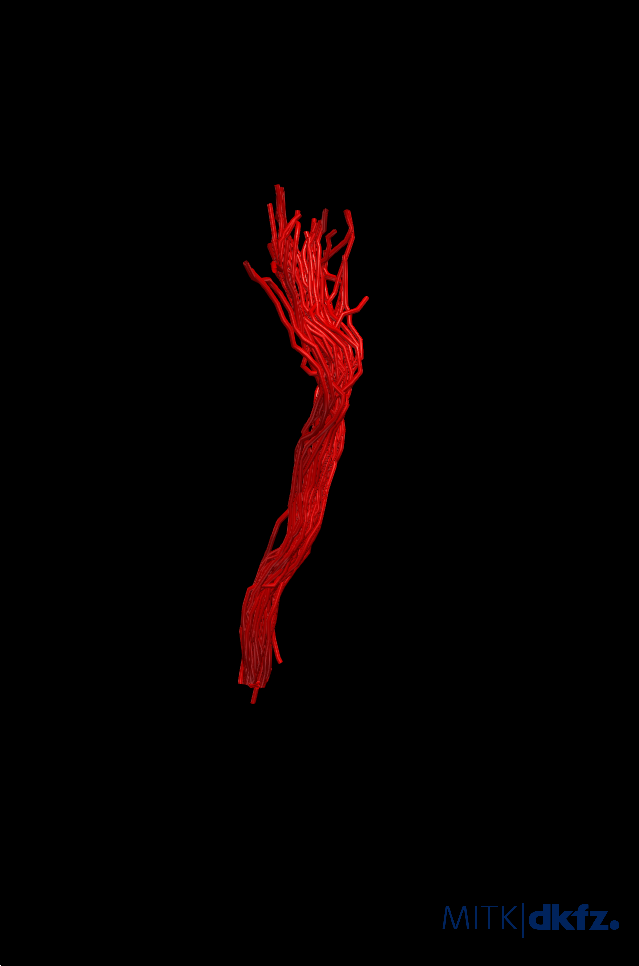}};
        \node[right=of pppn] (pppp) {\includegraphics[trim=1cm 5cm 1cm 4.5cm,clip,width=3.cm]{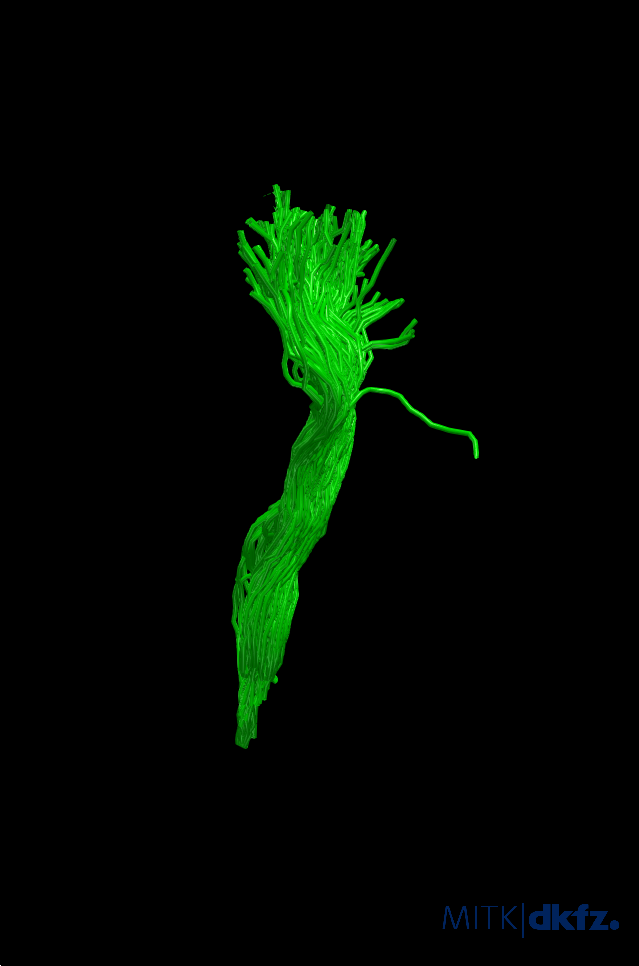}};
        %
        \tikzstyle{node label}=[below right=-7mm and -13mm of #1,text=white]
        \foreach \x in {nnnn, nnnp, nnpn, nnpp, npnn, npnp, nppn, nppp, pnnn, pnnp, pnpn, pnpp, ppnn, ppnp, pppn, pppp}
            \node[node label=\x] {\textsf{\x}};
    \end{tikzpicture}
    \caption{Visualisations of streamlines of each combination of individual supervisor labels (TQ, RBX, TS, AIF) for the right cortico-spinal tract. The streamlines are colored in red, green and blue to show that they will be  classified as positive, negative or inconclusive, respectively. Notice that some combinations do not contain streamlines.}
    \label{fig:CST-visualisations}
\end{figure}

\paragraph{Stability of the supervisors across subjects} We analysed the labels obtained from each of the supervisor methods individually. In Fig.~\ref{fig:supervisor_stats_4}, we report the percentage of streamlines for the positive and negative label obtained from each of the supervisor methods separately, averaged over all 20 subjects in our dataset. The following observations can be made from this figure. On the one hand, different supervisor methods result in a different ratio of positive to negative labels. While AIF results in mostly positive labels, the others produce mostly negative labels. On the other hand, The reported standard deviation shows that for AIF, TractQuerier, and RecobundlesX, the number of obtained streamlines is stable over subjects. However, in the case of TractSeg, we see differences of up to 14\% of the streamline count between certain subjects. 
\paragraph{Stability of the composition of individual supervisor labels across subjects} We analysed the subsets of streamlines obtained from combining the individual labels of all supervisor methods, resulting in sixteen different classes (cf. Sect.~\ref{sssec:three_class_problem}). The bar chart on the left in Fig.~\ref{fig:label_balance_4} shows that some individual classes are much more common than others in the tractogram. Further, the indicated standard deviations for each class show a higher variability for some classes across the subjects. Regarding the further aggregation of the composition labels according to Table~\ref{tab:three_class_problem} (cf.  Fig.~\ref{fig:label_balance_4} on the right) are the three classes \emph{POS}, \emph{U} and \emph{NEG}.  The three classes are sufficiently balanced (\emph{POS}: $18\%$, \emph{NEG}: $32\%$, \emph{U}: $50\%$) enabling us to train a classifier replicating this data separation.
\paragraph{Interpretation of the sixteen classes} When combining the individual supervisor methods, we do not expect it to appear with the same likelihood in the data. Focusing on the classes listed as borderline in Table~\ref{tab:composition_classes} (marked with a star in the bar plot of Fig.~\ref{fig:label_balance_4}), we observe that these show a very low occurrence compared to the other classes. The fact that these classes make up a total of $5.6\%$ of the overall data supports our hypothesis that these combinations of supervisor labels are caused by inaccuracies and are expected to have a minor impact on training the classifier.
\paragraph{Visualisation of the classes}
Figure \ref{fig:CST-visualisations} shows plots of the individual sixteen classes for the right cortico-spinal tract (CST). Red, green and blue are used to show the positive, negative and unknown/inconclusive classes, respectively. TractQuerier, RecobundlesX, and TractSeg compute positives per bundle, which is advantageous for visualisation of all classes in the right CST, except for nnnn and nnnp. For these two classes, we select streamlines that start and end in masks created by TractSeg and remove streamlines crossing the left hemisphere. Notice that the borderline classes \emph{npnn} and \emph{ppnn} contain more streamlines than in other bundles (cf. Fig. \ref{fig:label_balance_4}), which might be a characteristic of RecobundlesX for the CST.
\par
The inconclusive classes contain a mixture of plausible and implausible streamlines. A similar trend can be seen for other fibre bundles.
\subsection{Performance of the model}
\begin{figure}
    \centering
    \includegraphics[width=\textwidth]{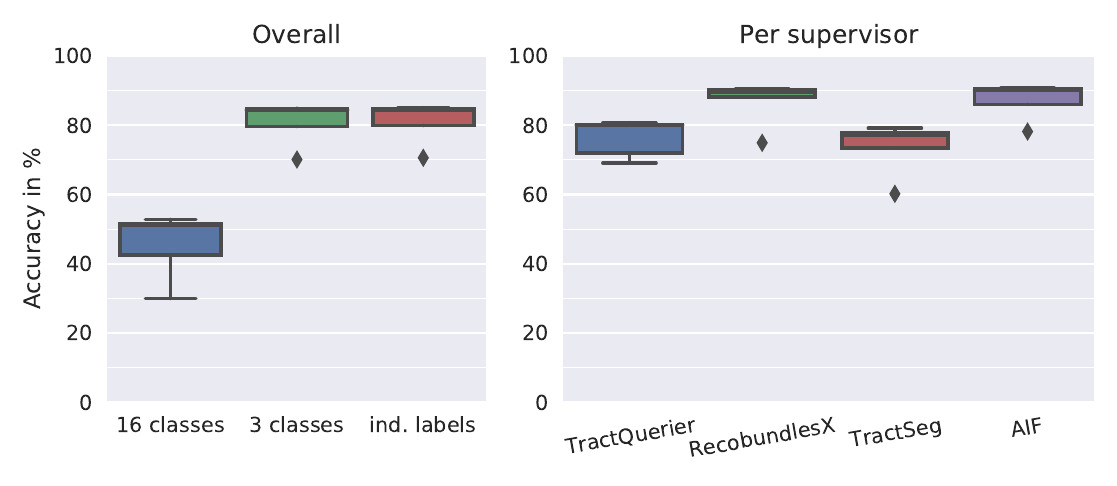}
    \caption{Achieved accuracy for the chosen model architecture. The trained model is evaluated on 500k samples drawn from the three evaluation subjects of a particular data split. \textbf{Left:} Accuracy is computed based on different sets of labels on a common set of data samples. The reported values relate to the composition classes from Table~\ref{tab:composition_classes} (sixteen classes), and Table~\ref{tab:three_class_problem} (three classes) and the average accuracy of the individual supervisor labels. \textbf{Right:} Accuracy for each of the individual supervisor labels.}
    \label{fig:model_acc}
\end{figure}
\begin{figure}
    \centering
    \includegraphics[width=\textwidth,clip,trim=0 0 0 0.8cm]{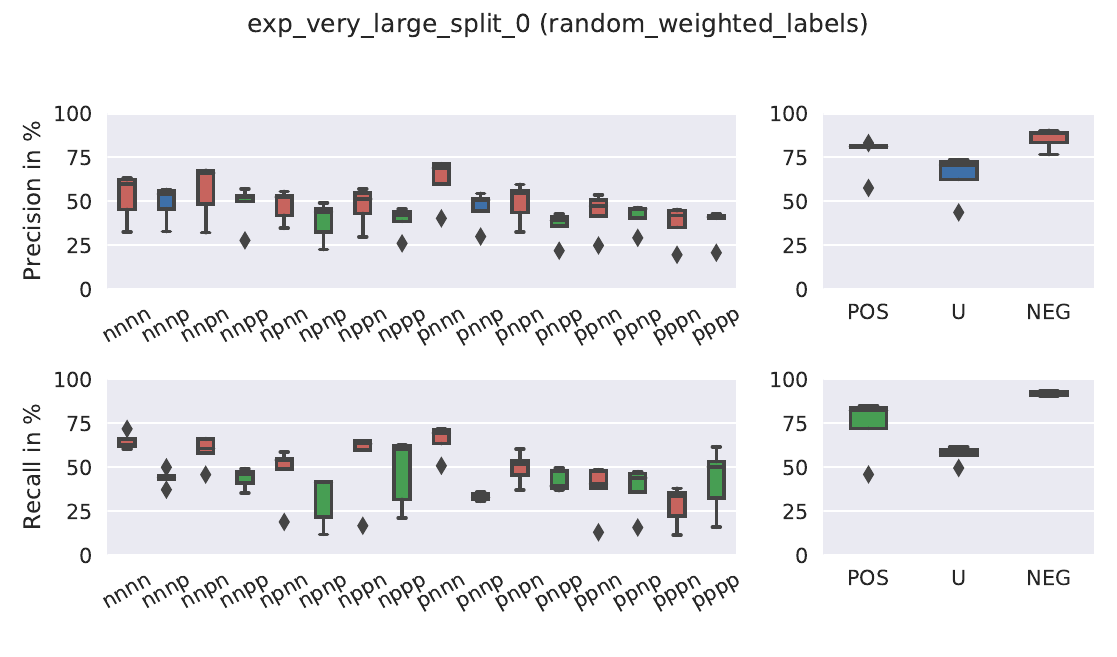}
    \caption{Precision and recall of the classification evaluated on 500k samples drawn from the three evaluation subjects of a particular data split. \textbf{Left:} Values are reported for the sixteen combinations of supervisor labels from Table \ref{tab:composition_classes}. \textbf{Right:} Values are reported for the three classes introduced in Table~\ref{tab:three_class_problem}.}
    \label{fig:model_pr}
\end{figure}
\paragraph{Prediction of individual supervisor labels} The accuracy achieved in predicting the individual supervisor label varies depending on the supervising method, as shown in Fig. \ref{fig:model_acc} on the right. For the labels of RecobundlesX and AIF the model achieves an accuracy between 85--90\%. This is a reasonable performance, considering the manual tuning of hyperparameters. However, the labels of TractQuerier and TractSeg are correctly predicted for about $75\%$ of the samples. This underlines that predicting these two classes is a more difficult task. 
\paragraph{Prediction of aggregated classes} We assessed the performance of the model based on 500k streamlines sampled from three evaluation subjects. These streamlines are selected based on the sampling strategy introduced in Subsect.~\ref{ssec:model_training}. As shown in Fig.~\ref{fig:model_acc} on the left, the mean accuracy of the prediction of the sixteen composition classes is 43\%. However, we achieve a reasonable accuracy of about $80\%$ for predicting the three classes: POS / NEG / U. This value is close to the average performance achieved for each of the individual supervisor methods (cf. Fig.~\ref{fig:model_acc} on the right).
\par
Figure~\ref{fig:model_pr} shows precision  and recall measurements. For the sixteen classes, we found that both measures show much variation among the individual classes. On the other hand, it can be noticed that the classes POS and NEG show relatively high precision and recall, which is an indication that  streamlines grouped in one of these two classes are very similar even if belonging to different of the sixteen composition classes. That would explain the lower accuracy for the sixteen-class problem and would also underline the meaningfulness of the grouping into three classes as introduced in Subsect.~\ref{sssec:three_class_problem}.
\subsection{Ablation study}
\begin{figure}
    \centering
    \includegraphics[width=\textwidth]{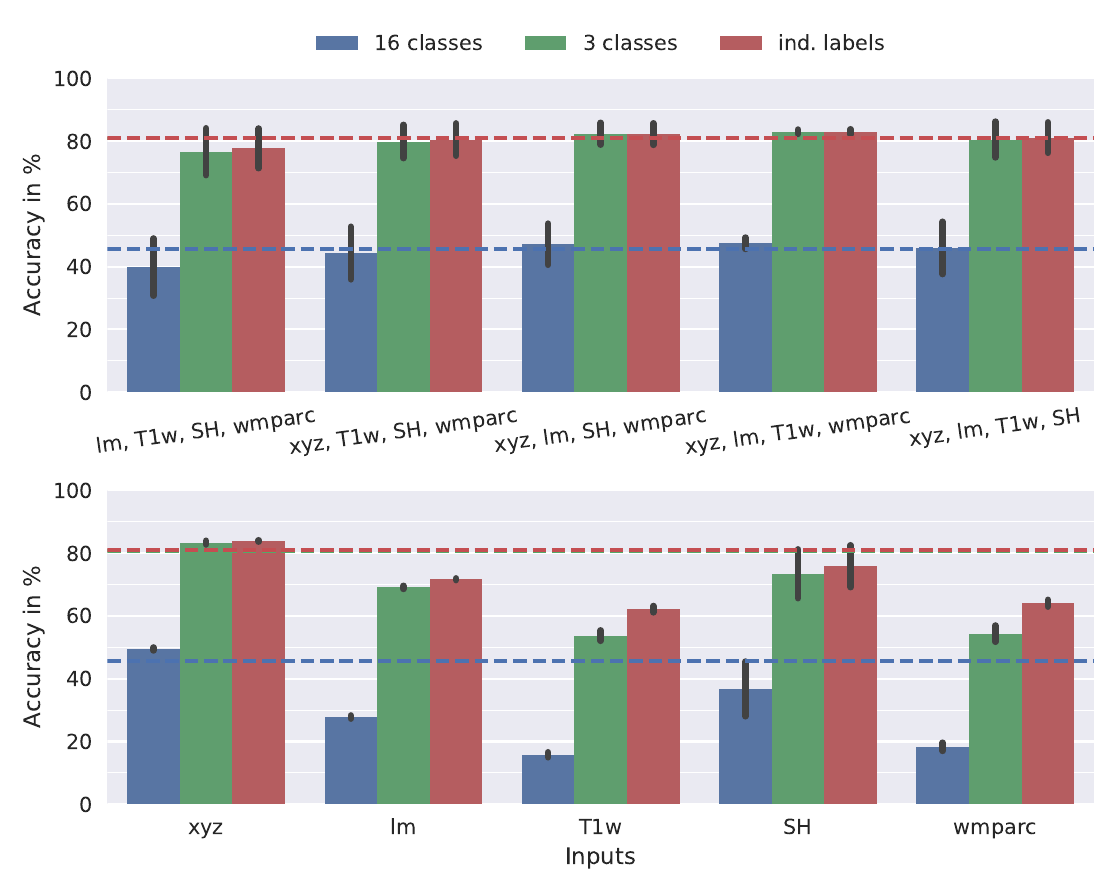}
    \caption{Accuracy achieved with different inputs. Accuracy for sixteen classes, three classes, and individual supervisor labels are reported for the same model architecture and different sets of inputs. Each model is trained based on the data of the mentioned inputs while inserting standard normal distributed noise into the branches of the others. Dashed lines indicate the performance of the model using all inputs. \textbf{Top row:} Training with \textit{all but one} input feature. \textbf{Bottom row:} Training with \textit{only a single} input. Inputs: \textbf{xyz}: streamline coordinates, \textbf{lm}: landmarks, \textbf{SH}: spherical harmonics of the dMRI data, \textbf{T1w}: T1w data, \textbf{wmparc}: parcellation from Freesurfer.} 
    \label{fig:nullify_overall}
\end{figure}
\begin{figure}
    \centering
    \includegraphics[width=\textwidth]{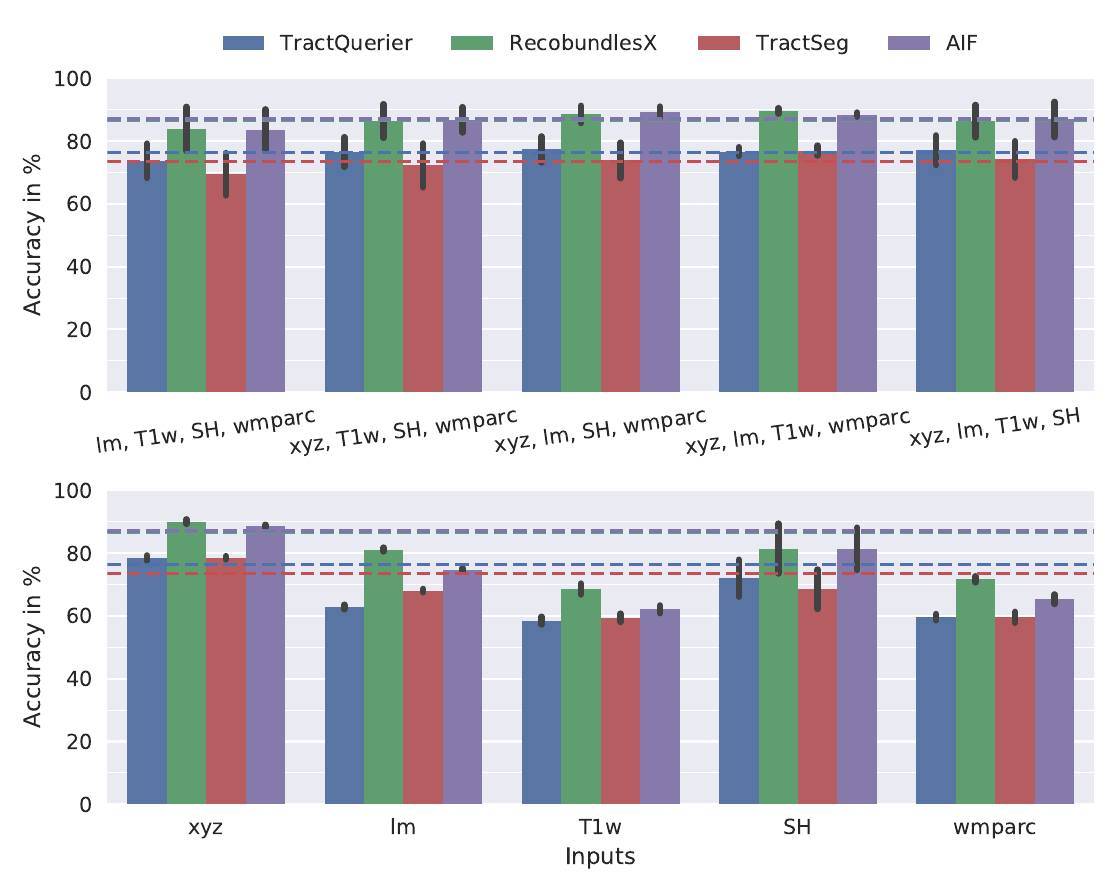}
    \caption{Accuracy achieved with different inputs for the four supervisors. Each model is trained based on the data of the mentioned inputs while inserting standard normal distributed noise into the branches of the others. Dashed lines indicate the performance of the model using all inputs. \textbf{Top row:} Training with \textit{all but one} input feature. \textbf{Bottom row:} Training with \textit{only a single} input. Inputs: \textbf{xyz}: streamline coordinates, \textbf{lm}: landmarks, \textbf{SH}: spherical harmonics of the dMRI data, \textbf{T1w}: T1w data, \textbf{wmparc}: parcellation from Freesurfer.}
    \label{fig:nullify_individual}
\end{figure}

\paragraph{Leaving an input feature out} Comparing the achieved accuracy of the retrained model when replacing the data of one input feature with random values, we found that only in the case of the streamline coordinates (`\textit{xyz}' in Fig. \ref{fig:method_overview}) the performance drops notably (see top rows of Figures~\ref{fig:nullify_overall} and \ref{fig:nullify_individual}). This indicates that there is information contained in the coordinates that the model cannot extract from the other inputs. In turn, almost the same accuracy is achieved when dropping the information of any of the other inputs. A possible explanation for this observation is that the dropped information seems to be redundant and is also present in one of the respective other inputs. However, since the accuracy always dropped after removing one of them, this experiment also shows that all inputs convey some relevant information for the classification.
\par
\paragraph{Using a single input feature} Following the reversed strategy, we trained the model using the information of only a single input at the time. As shown in the bottom rows of Figures~\ref{fig:nullify_overall} and \ref{fig:nullify_individual}, the model using solely streamline coordinates as input attains almost the same performance as compared to the model that uses all inputs. This finding is in line with the observations in the previous paragraph. 
\par
Regarding the other four inputs, dMRI is the second-best input followed by landmarks, the parcellation, and T1w, respectively. Interestingly, when focusing on the accuracy for each of the supervisor methods individually, we see that the use of diffusion data was especially boosting the results for AIF and RecobundlesX as compared to landmarks, T1w information, and the parcellation. As opposed to that, for TractQuerier and TractSeg the improvements due to the use of diffusion data are only mild.
\section{Discussion}
\label{sec:discussion}
As already mentioned, this paper has three major contributions. First, we proposed a methodology for separating streamlines into three disjoint classes: \emph{positive}, \emph{negative}, and \emph{inconclusive}. To the best of our knowledge, the new class \emph{inconclusive} has not been proposed before. Specifically, this new class accounted for approximately half of the streamlines in our experiments (cf. Fig. \ref{fig:label_balance_4}). Targeting the inconclusive class in new methods can be used to improve the filtering performance. Second, we proposed a methodology for combining supervisors based on their specific characteristics (cf. Tables \ref{tab:supervisors}, \ref{tab:composition_classes}, and \ref{tab:three_class_problem}). Using this methodology, we were able to train a neural network for reproducing both each supervisor individually as well as the proposed combination with an accuracy of about 80\% (cf. Fig. \ref{fig:model_acc}). Finally, we performed ablation experiments to assess the importance of different input descriptors in the classification of streamlines. Our experiments showed that the 3D coordinates of the streamlines are by far the most important ones, followed by the diffusion data along the streamline. In turn, T1w information, our proposed landmark descriptor, and a brain parcellation were less relevant for the classification task (cf. Fig. \ref{fig:nullify_overall} and \ref{fig:nullify_individual}). To our knowledge, this type of analysis has not been performed before.
%
\par
In the following subsections, we discuss the results in detail, point out the limitations of our approach and discuss issues to be addressed in future works.
\subsection{Label creation}
\label{ssec:disc_label_creation}
\paragraph{Choice of supervising methods} Although we used four supervisors in this paper, it is possible to use the same methodology for more in the future. One of the main advantages of adding more supervisors is that the size of the unknown/inconclusive class could very well be reduced, which is currently around 50\%. Specifically, Tables \ref{tab:composition_classes} and \ref{tab:three_class_problem} must be redesigned when adding new supervisors. Similarities between the new supervisors with the current ones can be used for the redefinition of classes. For example, one can add more bundle segmentation-based methods (e.g., \textit{Deep White Matter Analysis} \citep{Zhang2019}, or other bundle definitions in TractQuerier or RecobundlesX) and use a similar analysis to the one used for RecobundlesX or TractSeg. AIF can also be used as a reference for methods trying to target implausible streamlines (e.g., at some extent, the methods by \citet{Astolfi2020TractogramLearning}, \citet{Legarreta2021FilteringFINTA} or \citet{Hain2023}). It is also important to notice that the  current definition of anatomical bundles is largely dependent on the virtual dissection pipeline \citep{Schilling2021TractographyDataset,Rheault2020Tractostorm:Reproducibility}. This issue must be taken into account since it can have a large impact in the filtering results.
\par
Adding supervisor methods based on data fitting such as COMMIT \citep{Daducci2015}, COMMIT2 \citep{Schiavi2020AInformation,Ocampo-Pineda2021HierarchicalTractography}, SIFT \citep{Smith2013}, SIFT2 \citep{Smith2015a} or LiFE \citep{Pestilli2014} is appealing due to their direct connection to the measured diffusion signal. Unfortunately, important issues must be solved before these methods can be added to the model. Considering for example SIFT, a natural definition of the positive label could be to ascribe it to all streamlines obtained from a tractogram after application of this method, and negative to the rest. However, a streamline with a positive label does not necessarily reflect anatomy, so it can actually be implausible. On the other hand, the negative label also includes plausible but redundant streamlines \citep{Jorgens2021ChallengesFiltering}. These issues make it difficult to incorporate such a method as a supervisor in Tables \ref{tab:composition_classes} and \ref{tab:three_class_problem} to distinguish between anatomically plausible and implausible streamlines. \citet{Hain2023} and \citet{Wan2023} tackled this problem by running SIFT and COMMIT several times, respectively.
\paragraph{Definition of classes} While the division of streamlines into three classes is natural to argue for, the question of how to define it requires a more careful argumentation. The mapping in Table~\ref{tab:three_class_problem} is based on the interpretation of all possible combinations of the used supervisor methods provided in Table~\ref{tab:composition_classes}. As discussed above, it is clear that a similar argumentation must be performed when including more supervisor methods. 
\par
We argued in Subsect.~\ref{sssec:three_class_problem} that the occurrence of the class labels grouped as borderline in Table~\ref{tab:composition_classes} should be minor. As shown in Fig.~\ref{fig:label_balance_4}, these seven labels are among those with the lowest numbers. That supports the assumption that these cases are due to inaccuracy in masking operations and due to the low fraction of $5.6\%$ should not have a major effect on the classification with the three classes. 
\par
Notice that the reported model performance in terms of accuracy is close to the accuracy of the individual supervisor predictions. This indicates that the mapping in Table~\ref{tab:composition_classes} is meaningful. When predicting the individual sixteen class labels, the accuracy of the model is low. However, grouping several of those classes together captures wrong classifications in the same group resulting in higher accuracy. We hypothesise that streamlines grouped in either class POS or NEG might be very similar within each class. However, they might be grouped into different classes when considering the sixteen supervisor combinations. The observed accuracies support this hypothesis.
\paragraph{Variability in TractSeg results} The fractions of streamlines with a \textit{positive} or a \textit{negative} label obtained from  TractSeg showed a larger subject variability than for the other supervisor methods (cf. Fig.~\ref{fig:supervisor_stats_4}). The standard deviation was up to 2.5 times larger than for the other methods.  To better understand the reason for this, we applied the TractSeg supervisor to tractograms computed without the option of \textit{anatomically constrained tractograpy} (ACT) (as described in \cite{Wasserthal2018}). In this case, the standard deviation was lower than for the other supervisors, which means that the main source of variability of this supervisor was the use of ACT for creating the tractogram. This highlights the individual subject's anatomy as a potential cause for this increased variability. An additional question is whether or not such variability of TractSeg is bundle-dependent, which was not the case. This strengthens our assumption that the more significant variability in the TractSeg supervisor stems from individual subject variations rather than inaccurately computed masks by TractSeg.
\paragraph{Variability in the composition classes} The coefficient of variation for the individual composition classes is relatively high for some of the classes (cf. Fig.~\ref{fig:label_balance_4} on the left). One possible reason for this is that the variability of TractSeg might be propagated when combining it with the other supervisor labels. On the other hand, other inter-dependencies between the individual supervisor labels can potentially be a source of variability. The fact that the accuracy of the trained models is high for AIF and RecobundlesX, which have low variability in their label fractions, but that it is low for TractSeg--having a high label variability--and TractQuerier--having a low label certainty--might hint to a relation of the stability of the created labels, and the ability of the model to learn their mapping from the data.
\subsection{Data-driven model}
\paragraph{Model performance} The main goal of this paper was to show the feasibility of training a model that is able to replicate the prediction of the three-class problem described in Subsection \ref{sssec:three_class_problem}, in order to leverage the computational efficiency and ease of use provided by a machine learning model in inference mode. Our results showing an accuracy of around $80\%$ are a promising indication that the stated goal is achievable. Due to the empirical tuning of hyperparameters applied so far, we expect further improvements in performance with a more structured optimisation approach. 
\paragraph{Prediction of individual supervisor labels} As already mentioned, the performance for predicting the individual labels varies substantially among the supervisor methods. It is apparent that the prediction of the RecobundlesX label is relatively easy, which is also reflected by a very early convergence of the learning curve after a few epochs during training. Furthermore, AIF was also relatively easy to learn, although its learning curve was flatter. On the other hand, the performance for TractSeg and TractQuerier was less satisfactory. We hypothesise that the performance of both is related to the characteristics of their labels, as mentioned above. In the case of TractSeg, we observed that the test performance was lower than in training. This shows that the variability across subjects is important and might require more  training data if the goal is to replicate TractSeg.  For TractQuerier, the low certainty of the respective binary labels is likely to be the reason for the difficulty of achieving a better prediction.
\paragraph{Improvement of training performance} In Table~\ref{tab:composition_classes}, we introduced a group of labels as borderline cases. The knowledge of these could---if directly encoded in the network---be helpful for the performance. Voting / attention-like connections between the final classifiers in the different output branches could help the model avoid predicting the borderline cases. That could be beneficial for the prediction of the three classes. Alternatively, the three classes could be used to define the loss, which is optimised in training. While this could help to overcome inaccuracy in the supervising methods and potentially result in better performance, it would come at the cost of interpretability if the supervisor labels would not be predicted individually.
\paragraph{Generalisability of the model} In the experiments, the performance of the proposed model was only tested on one set of tractography data and the same measurement data (HCP). Our experiments showed that the coordinates of the streamlines were the most important input for classification. It would be interesting to assess if that result is also valid for other datasets, tractography methods, and their parameters. 
These experiments are beyond the scope of this paper but will be reported in subsequent publications. 
\subsection{Sensitivity to the input data}
\paragraph{Predictive value of streamline coordinates} When restricted to a single input, the model achieved the highest accuracy with information from streamline coordinates. The final performance was almost as good as when all inputs were used. On the other hand, leaving out only the streamline coordinates led to an important performance reduction of the model compared to any other input feature. These two observations reflect the relevance of streamline coordinates for classification. Streamlines are defined solely by the sequence of their coordinates, and therefore a mapping from that set of information is expected to perform reasonably well. On the other hand, another reason for these observations might be the nature of the four employed supervising methods, all of which are based on the geometry of the streamlines, that is, information that is encoded in the coordinates. That means that adding methods strongly based on the diffusion data, like COMMIT or SIFT, might alter this dependence. 
From the sensitivity to other inputs shown in Figures~\ref{fig:nullify_overall} and \ref{fig:nullify_individual} it is apparent that the model manages to extract predictive information from the other inputs, especially from the diffusion data. Therefore, if another type of supervising method would require information that is not contained in the streamline coordinates alone, the other inputs might be more important for the label prediction. Notice that we just  normalise the coordinate values to the range [-1,1]. It is to be seen whether pre-aligning the tractograms to the MNI template can be used for improving the accuracy of the neural network, something we shall explore in future works.
\paragraph{Assessing the value of an individual input feature} In our experiments, we aimed at assessing the value of individual inputs through an ablation study. The reported accuracy for models trained only on information from one streamline feature estimates the information included in the data for our predictions. However, the second series in which we assess the performance when training with \textit{all but one} input feature results in less contrast to distinguish the trained models. While we can clearly see that the information from the streamline coordinates cannot be obtained from the other four inputs (cf. Fig.~\ref{fig:nullify_overall}, top row), we cannot see any difference when leaving out any of the other inputs. Testing further reduced sets of inputs could reveal more dependencies between them.
\par
Another interesting observation is that when training on all but the streamline coordinates, the performance was better than with any other input considered individually. This probably stems from the fact that each of the models trained on the individual inputs showed a slightly different performance in the individual supervisor prediction (cf. Fig.~\ref{fig:nullify_individual}, bottom row). We conclude from this that there is a value of combining different input descriptors for training the model that might be advantageous when aiming to improve the performance for all supervisors and potentially when including further supervisors.
\section{Conclusion}
One way of improving the specificity of tractography for exploring the anatomy of the human brain is to apply careful post-processing of the obtained tractograms. In this paper, we propose a novel approach for streamline filtering that can divide a tractogram into three disjoint sets of streamlines representing anatomically plausible, implausible, and unknown/inconclusive streamlines. We achieve this by combining the knowledge from four strategies for streamline classification. Our experiments show that the separation of streamlines into the three classes can be achieved with a well-trained neural network with high accuracy. Finally, we found out that the 3D coordinates of the streamlines, followed by the diffusion data along the streamline, are the most informative descriptors for tractogram filtering with respect to the selected supervisors.
%
%
%
%
%
%
%
\section*{Acknowledgements}
We thank Jakob Wasserthal for computing and providing the tractograms we used in this paper.
\section*{Funding sources}
This work was partially supported by VINNOVA, through AIDA; Digital Futures, project dBrain and The Swedish Research Council grant No. 2022-03389. Pierre-Marc Jodoin has been funded by the Canadian NSERC Discovery grant No.03SNG008. Part of this work was also possible thanks to the Université de Sherbrooke research chair in Neuroinformatics and NSERC Discovery grant RGPIN-2020-04818. The funding sources had no involvement in the research and preparation of this article.
\section*{Author contributions}
\textbf{DJ}: conceptualization; formal analysis; investigation; methodology; software; validation; visualization; writing - original draft; writing - review \& editing. 
\textbf{PMJ}: conceptualization; funding acquisition; methodology; resources; supervision; writing - review \& editing. 
\textbf{MD}: conceptualization; funding acquisition; methodology; resources; supervision; writing - review \& editing. 
\textbf{RM}: conceptualization; funding acquisition; methodology; visualization; resources; project administration; supervision; writing - review \& editing. 
\section*{Conflicts of interest/Competing interests}
The authors declare that they have no conflict of interest.
\section*{Data and code availability statement}
We used data from the Human Connectome Project, WU-Minn Consortium (Principal Investigators: David Van Essen and Kamil Ugurbil; 1U54MH091657) funded by the 16 NIH Institutes and Centers that support the NIH Blueprint for Neuroscience Research; and by the McDonnell Center for Systems Neuroscience at Washington University.
%
%
%
%

%
\appendix
\section{Neural network architecture}
\label{app:network}
\begin{table}[ht]
    \caption{Specification of convolutional blocks in input branches.}
    \label{tab:input_branches}
    \centering
    \begin{tabular}{@{}llllll@{}}
        \toprule
        Input branch & xyz & lm & T1w & SH & wmparc \\
        \midrule
        Nb. of conv. blocks & 12 & 12 & 12 & 12 & 12 \\
        Nb. of kernels per conv. block & 208 & 208 & 208 & 416 & 208 \\
        Kernel size & 3 & 3 & 3 & 3 & 3 \\
        Block ids with pooling (sz: 2) & 4, 8 & 4, 8 & 4, 8 & 4, 8 & 4, 8\\
        \bottomrule
    \end{tabular}
\end{table}
\begin{table}[ht]
    \centering
    \caption{Specification of convolutional blocks in output branches.}
    \label{tab:output_branches}
    \begin{tabular}{@{}llllll@{}}
        \toprule
        Output branch & TQ & RBX & TS & AIF \\
        \midrule
        Nb. of conv. blocks & 4 & 4 & 4 & 4 \\
        Nb. of kernels per conv. block & 208 & 208 & 208 & 208 \\
        Kernel size & 5 & 5 & 5 & 5 \\
        Block ids with pooling & N/A & N/A & N/A & N/A \\
        \bottomrule
    \end{tabular}
\end{table}
\begin{table}[ht]
    \centering
    \caption{Specification of the final series of fully connected (FC) layers in each output branch. Note that the input to the first FC layer is the flattened output of the last convolutional block of each output branch. Further note that the input to the third FC layer is obtained through concatenation of the outputs of the second FC layer in all four output branches.}
    \label{tab:output_branches2}
    \begin{tabular}{@{}llllll@{}}
        \toprule
        Layer & Input size & Output size \\
        \midrule
        FC 1 & 10400 & 196 \\
        FC 2 & 196 & 196 \\
        FC 3 & 784 & 196 \\
        FC 4 & 196 & 2 \\
        \bottomrule
    \end{tabular}
\end{table}
\end{document}